\documentclass{article}

\usepackage{microtype}
\usepackage{graphicx}
\usepackage{subfigure}
\usepackage{booktabs} 

\usepackage{hyperref}

\usepackage[accepted]{icml2025}

\usepackage{amsmath}
\usepackage{amssymb}
\usepackage{mathtools}
\usepackage{amsthm}

\usepackage[capitalize,noabbrev]{cleveref}

\theoremstyle{plain}

\theoremstyle{definition}

\theoremstyle{remark}

\usepackage[colorinlistoftodos,bordercolor=orange,backgroundcolor=orange!20,linecolor=orange,textsize=scriptsize]{todonotes}

\newcommand{\del}[1]{}

\newcommand{\eqdef}{\stackrel{\text{def}}{=}}

\newcommand{\RD}{\mathbb{R}^d}

\definecolor{linenumbercolor}{rgb}{0.98, 0.81, 0.69}

\definecolor{classcolor}{RGB}{0,0,255}

\definecolor{apicolor}{RGB}{255,0,0}

\definecolor{linenumbercolor}{rgb}{0.1, 0.1, 0.1}

\definecolor{modelcolor}{RGB}{108,57,0}
\definecolor{datacolor}{RGB}{108,57,0}
\definecolor{abrcolor}{RGB}{135,106,181}
\definecolor{algcolor}{RGB}{108,57,0}
\definecolor{compcolor}{RGB}{108,57,0}
\definecolor{libcolor}{RGB}{108,57,0}

\newcommand{\modelname}[1]{{\color{modelcolor}\sf \small {#1}}} 

\newcommand{\algname}[1]{{\color{algcolor}\sf \small {#1}}}     
\newcommand{\compname}[1]{{\color{compcolor}\sf \small {#1}}}   
\newcommand{\libname}[1]{{\sf \color{libcolor} \small {#1}}}    

\usepackage{csquotes}
\usepackage{longtable}
\usepackage[flushleft]{threeparttable}
\usepackage{colortbl}
\usepackage{xcolor}
\usepackage{makecell}
\usepackage{nicefrac}
\usepackage{enumitem}
\usepackage{microtype}
\usepackage{graphicx}
\usepackage{subfigure}
\usepackage{multicol}
\usepackage{listings}
\usepackage{natbib}

\usepackage{pgfplots}
\usepackage{tikz}

\definecolor{bgcolorwe}{rgb}{1.0,1.0,0.6}
\definecolor{abscolor}{rgb}{0.501,0.521,0.533}
\definecolor{myblue}{rgb}{0.1, 0.3, 0.7}

\lstdefinestyle{mystyle}{
    backgroundcolor=\color{white},   
    commentstyle=\color{darkgray},      
    keywordstyle=\color{magenta},    
    numberstyle=\tiny\color{gray},   
    stringstyle=\color{purple},      
    basicstyle=\ttfamily\footnotesize, 
    breakatwhitespace=false,         
    breaklines=true,                 
    captionpos=t,                    
    numbers=left,                    
    numbersep=5pt,                   
    showstringspaces=false           
}
\icmltitlerunning{BurTorch}

\begin{document}
\onecolumn
\icmltitle{BurTorch: Revisiting Training from First Principles by Coupling Autodiff, Math Optimization, and Systems}




\icmlsetsymbol{equal}{*}

\begin{icmlauthorlist}
\icmlauthor{Konstantin Burlachenko}{yyy}
\icmlauthor{Peter Richt\'{a}rik}{yyy}
\end{icmlauthorlist}

\icmlaffiliation{yyy}{King Abdullah University of Science and Technology, Kingdom of Saudi Arabia}

\icmlcorrespondingauthor{Konstantin Burlachenko}{konstantin.burlachenko@kaust.edu.sa}
\icmlcorrespondingauthor{Peter Richt\'{a}rik}{peter.richtarik@kaust.edu.sa}

\icmlkeywords{optimization software, federated learning, high-performance implementation, backpropagation, automatic differentiation, C++}

\vskip 0.3in


\printAffiliationsAndNotice{}  

\begin{figure}[ht]
    \centering
    \includegraphics[width=105px, keepaspectratio=true]{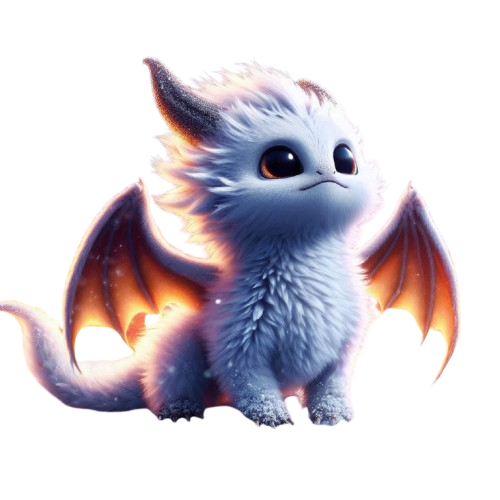}
\end{figure}

\begin{abstract}
\label{abs}
\setcounter{footnote}{1}

In this work, we introduce \libname{BurTorch}\footnote{
The BurTorch (Backpropagation Ultrafast Runtime) logo is a DALL-E 3 modified version of the \href{https://github.com/karpathy/micrograd/blob/master/puppy.jpg}{Micrograd logo} \cite{karpathy2020micrograd}.}, a compact high-performance framework designed to optimize Deep Learning (DL) training on single-node workstations through an exceptionally efficient CPU-based backpropagation \citep{rumelhart1986learning, linnainmaa1970representation} implementation. Although modern DL frameworks rely on compiler-like optimizations internally, \libname{BurTorch} takes a different path. It adopts a minimalist design and demonstrates that, in these circumstances, classical compiled programming languages can play a significant role in DL research. By eliminating the overhead of large frameworks and making efficient implementation choices, \libname{BurTorch} achieves orders-of-magnitude improvements in performance and memory efficiency when computing $\nabla f(x)$ on a CPU. \libname{BurTorch} features a compact codebase designed to achieve two key goals simultaneously. First, it provides a user experience similar to script-based programming environments. Second, it dramatically minimizes runtime overheads. In large DL frameworks, the primary source of memory overhead for relatively small computation graphs $f(x)$ is due to feature-heavy implementations. We benchmarked \libname{BurTorch} against widely used DL frameworks in their execution modes: \libname{JAX} \citep{jax2018github}, \libname{PyTorch} \citep{paszke2019pytorch}, \libname{TensorFlow} \citep{abadi2016tensorflow}; and several standalone libraries: \libname{Autograd} \citep{maclaurin2015autograd}, \libname{Micrograd} \citep{karpathy2020micrograd}, \libname{Apple MLX} \citep{mlx2023}. For small compute graphs, \libname{BurTorch} outperforms best-practice solutions by up to $\times 2000$ in runtime and reduces memory consumption by up to $\times 3500$. For a miniaturized \modelname{GPT-3} model \citep{gpt}, \libname{BurTorch} achieves up to a $\times 20$ speedup and reduces memory up to $\times 80$ compared to \libname{PyTorch}.

\end{abstract}

\section{Introduction}
\label{sec:intro}

First-order optimization methods are a class of continuous optimization techniques that rely on gradient and function value information to minimize a target function, without using higher-order derivatives. For instance, the Gradient Descent (\algname{GD}) algorithm iteratively updates model parameters to approach a stationary point. Specifically, \algname{GD} updates the model parameters using the rule:
\[
x^{k+1} \eqdef x^k - \gamma \nabla f(x^k),
\]
where $\gamma \in \mathbb{R}$ is the learning rate, and $\nabla f(x^k) \in \mathbb{R}^d$ is the gradient of the objective at the current iterate $x^k$. If $\nabla f(x^k)$ does not exhibit any structure, this model update for \algname{GD} requires $d$ in-place scalar additions and multiplication by $\gamma$. The time complexities of most optimization methods, including theoretically optimal algorithms such as \algname{Nesterov Accelerated Gradient Descent} for convex objectives \citep{nesterov2018lectures} and \algname{PAGE} \citep{li2021page,tyurin2022sharper} for cases where
\begin{equation}
	\label{eq:main}
	f(x) = \frac{1}{n} \sum_{i=1}^{n} f_i(x),
\end{equation}
and $f_i(x)$ is non-convex, are typically expressed in terms of gradient oracle complexities. For this reason, efficient gradient computation is crucial, as the cost of gradient evaluations significantly impacts the overall training cost. One way to estimate $f(x)$ from Equation~\ref{eq:main} in an unbiased manner is by sampling a subset $S \subseteq [n], |S|=b$ uniformly at random from all possible subsets of cardinality $b$, and constructing:
\begin{equation}
	\label{eq:main-est}
	f_{S}(x) = \dfrac{1}{b} \sum_{i \in S} f_{i}(x).
\end{equation}

For a differentiable function $f_S(x)$ from Equation~\eqref{eq:main-est}, the following holds due to the linearity of differentiation:
\begin{equation}
	\label{eq:main-grad-est}
	\nabla f_{S}(x) = \dfrac{1}{b} \sum_{i \in S} \nabla f_i(x). 
\end{equation}

Next, assume that each $f_i(x)$ for $i \in S$ is parameterized by a compact information description, such as an input-output pair in supervised machine learning problems. In this case, it is natural to assume that the \textit{time} to compute $\nabla f_{S}(x)$ in Equation~\ref{eq:main-grad-est} is equal to the sum of the times for computing $\nabla f_i(x)$ for $i \in S$.

However, this may not hold precisely in current practice. Modern Deep Learning frameworks such as \libname{JAX} \citep{jax2018github}, \libname{PyTorch} \citep{paszke2019pytorch}, and \libname{TensorFlow} \citep{abadi2016tensorflow} are the result of collaborative efforts across various disciplines. To effectively mask internal latencies, these frameworks employ a throughput-optimized design. The relatively large batch sizes $b$ help mask latencies by batching input and intermediate data across all software and hardware layers involved in the computation. This design, while optimized for throughput, does not provide an efficient method for computing individual gradient oracles $\nabla f_i(x)$. 

Furthermore, large frameworks comprising millions of lines of code, even when open-source, are often difficult to modify and computationally optimize. A computationally optimized implementation can follow two main philosophies: (a) perfecting every detail, and (b) limiting the number of details\footnote{Jack Dorsey is an engineer who applied this approach to create systems for society.}. However, as the implementation scales and requires human intervention, both philosophies are often compromised.

To address these challenges, we introduce \libname{BurTorch}, a minimalist framework that significantly improves both the latency of computing $\nabla f_i(x)$ and overall memory efficiency. Throughput-oriented Deep Learning (DL) frameworks incur high memory overhead during $\nabla f(x)$ execution (see Appendix~\ref{sec:backprop-memory} for the memory taxonomy within backpropagation), especially for large batch sizes, $b \gg 1$. \libname{BurTorch} mitigates this issue by computing individual gradient oracles $\nabla f_i(x)$ sequentially, reducing peak memory usage from $\sum_{i=1}^{b} \mathrm{MEM}(\nabla f_i(x))$ to $\max_{i \in [b]} \mathrm{MEM}(\nabla f_i(x))$.

\subsection{Mathematical tools for gradient computation} \label{sec:background-gd-compute}

One way to view the relation between $f(x): \mathbb{R}^d \to \mathbb{R}$ and its high-order derivatives is through the Taylor expansion at the point $x\in\mathbb{R}^d$ with the Lagrange remainder:

\begin{eqnarray}
	\label{eq:taylor-exp-with-2-terms}
	\dfrac{f(x + s \cdot \varepsilon) - f(x)}{\varepsilon} = \langle s, \nabla f(x) \rangle + \dfrac{s^\top \nabla^2 f(z) s \cdot \varepsilon}{2},
\end{eqnarray}
where $\varepsilon \in \mathbb{R}$, $s \in \mathbb{R}^d$, and $z \in (x, x + s \cdot \varepsilon)$.

\paragraph{Finite difference.}

If $\varepsilon \ll \min(1, \lambda_{\text{max}}(\nabla^2 f(z)))$, neglecting the second-order term in Equation~\eqref{eq:taylor-exp-with-2-terms} gives the \textit{forward finite difference method}, a local approximation of the directional derivative of $f(x)$ at $x$ in any direction $s$, $\|s\|_2=1$. From a computational standpoint, finite-difference methods face three challenges: (i) for small $\varepsilon$, numerical instability due to floating-point \citep{IEEE754-2008} limitations is not negligible; (ii) for large $\varepsilon$, error grows unbounded if $f(x)$ has non-zero curvature in direction $s$; (iii) finite-difference schemes require repeated evaluations of $f(x + s_i \cdot \varepsilon)$ for projections of $\nabla f(x)$ in multiple directions, leading to the overhead of $\times d$ for computing $\hat{\nabla f(x)} \approx \nabla f(x)$ using the numerical scheme \eqref{eq:taylor-exp-with-2-terms}.

\paragraph{Symbolic gradient oracles.}

When the structure of $\nabla f(x)$ is simple, gradient oracles can be derived symbolically and refined numerically. For example, \citet{burlachenko2024unlocking} explores methods to improve gradient and Hessian oracles for \modelname{logistic regression} symbolically. Methods that aim to manage computational graphs symbolically generally face two main challenges: (i) already combining $x_1, \dots, x_n \in \mathcal{X}$ with a single binary operator leads to a number of all possible distinct compute graphs asymptotically bounded by the Catalan numbers $\Omega \left( \nicefrac{4^n}{n^{3/2}} \right)$ \citep{cormen2022introduction}; (ii) the number of terms in symbolic gradient expressions can increase rapidly. For example, $\nabla_x \left( \prod_{i=1}^{n} x_i \right) = \left[ \prod_{j=1, j \neq i}^{n} x_j \right]_{1 \le i \le n}$ the symbolic gradient contains $(n^2 - n)$ terms, even though the original expression having only $n$ \citep{griewank2008evaluating}. Despite these complexities, there is a line of research that operates at this level~\citep{fawzi2022discovering, jia2019taso}.

\paragraph{Automatic Differentiation.} The \textit{Automatic Differentiation (AD)}, also known as \textit{Algorithmic Differentiation}, is an exact method for computing $\nabla f(x)$, assuming the exact execution of arithmetic and explicit description of $f(x)$. AD applies the chain rule systematically, unlike symbolic differentiation, by working with numerical values instead of symbolic expressions. The AD is fundamental in ML, where gradient computation is typically automated using the \textit{{Backpropagation Algorithm}} \citep{rumelhart1986learning}, a form of AD in reverse accumulation mode \citep{linnainmaa1970representation}. The theoretical computational cost and memory access cost of backpropagation can be conservatively estimated by the $w_{\text{tang}}$ cost of computing $f(x)$ and the cost of the memory accesses. The constant $w_{\text{tang}} \in [3, 4]$ is universal \citep{griewank2008evaluating}, and this bound holds if compute operations and memory accesses for $f(x)$ are in some sense atomic.

Backpropagation reuses intermediate computations to construct the necessary Jacobians mainly in two passes. The \textit{Forward Pass} performs inference and stores intermediate variables (activations), while the \textit{Backward Pass} implicitly computes the required Jacobians and generates $\nicefrac{\partial f(x)}{\partial x_i}$. Storing activations after the \textit{Forward Pass} is crucial for {backpropagation}. While {backpropagation} is preferred for computing the entire $\nabla f(x)$, {AD in Forward Accumulation Mode} \citep{rall1981automatic} is more memory-efficient when only a single directional derivative $\langle \nabla f(x), s \rangle$ is needed, as it computes the directional derivative in a single pass, mirroring the computation of $f(x)$. The cost of computing $f(x)$ and $\langle \nabla f(x), s \rangle$ with this method is bounded by a factor in the interval $[2, \nicefrac{5}{2}]$ relative to the cost of $f(x)$ evaluation \citep{griewank2008evaluating}.

\subsection{Progression of deep learning frameworks}
\label{sec:prev-systems}

Next, we review several key DL systems, with backpropagation playing a central role in their structure. See Appendix \ref{app:exex-details-of-comp} for details on eager and graph modes.

\paragraph{{Torch (2002).}} The modern era of Deep Learning frameworks began with Torch \citep{collobert2002torch}, which laid the foundation for many subsequent systems, including \libname{PyTorch} \citep{paszke2019pytorch}. \libname{Torch} was built around the Lua scripting language, which played a key role in its flexibility.

\paragraph{\color{black}{Theano (2010).}} \libname{Theano} \citep{bergstra2010theano} pioneered efficient Automatic Differentiation (AD) and optimization, accelerating the development of complex mathematical models. It introduced the computation graph abstraction, enabling automatic gradient computation, thereby streamlining the development of DL models. The graph-based approach in \libname{Theano} (and \libname{TensorFlow 1.0}) required careful graph construction before training.

\paragraph{\color{black}{Caffe 1.0 (2014).}} In 2014, \libname{Caffe} \citep{jia2014caffe} emerged as a leading framework for computer vision, simplifying model design through its \textit{Layer} abstraction and introducing runtime memory preallocation. Unlike \libname{Theano}, \libname{Caffe} did not require the same level of manual graph construction.


\paragraph{\color{black}{TensorFlow 1.0 (2015).}} In 2015, \libname{TensorFlow 1.0} \citep{abadi2016tensorflow} became the dominant framework in academia and industry. It functioned as both a programming language and a compiler, relying on an explicit computation graph managed by a \textit{Session} object. Key optimizations included automatic memory buffer reuse, constant folding and propagation, concurrent execution of independent operations, and memory allocation for variable states.


\paragraph{\color{black}{{PyTorch} (2016).}} From its inception, \libname{PyTorch} \citep{paszke2019pytorch} gained popularity for its intuitive debugging and flexibility in experimentation. Unlike \libname{TensorFlow 1.0}, which uses a static graph, \libname{PyTorch} employs an imperative computation style that integrates implicit graph construction with the forward pass. While this approach offers greater flexibility, it complicates optimization. The design of \libname{TensorFlow 2.0} mirrored \libname{PyTorch}'s eager execution model.



\paragraph{\color{black}{Keras (2015) and TensorFlow 2.0 (2019).}} \libname{TensorFlow 2} introduced \textit{eager execution}, enabling computations without the need for explicit graph construction, bringing it closer to the dynamic graph approach of \libname{PyTorch}. Although this improved flexibility, it came at the cost of performance. The benefits of eager execution include: (i) immediate error reporting, (ii) removal of legacy constructs such as placeholders and sessions, (iii) seamless integration with language-specific data structures and control flows, and (iv) compatibility with native debugging tools. \libname{Keras} \citep{chollet2015keras} was integrated into \libname{TensorFlow 2} as its API.


\paragraph{\color{black}{JAX (2018).}} \libname{JAX} \citep{jax2018github} extends \libname{NumPy} \citep{walt2011numpy} with AD and hardware acceleration. While its eager mode lacks efficiency, its just-in-time (JIT) compilation for static graph execution focuses on highly optimized internal computations.

\paragraph{\color{black}{Standalone packages.}} \libname{Autograd} \citep{maclaurin2015autograd}, \libname{Micrograd} \citep{karpathy2020micrograd}, and \libname{Apple MLX} \citep{mlx2023} focus solely on AD implementation.


\subsection{Motivation behind our work}
\label{sec:motivation}

The foundations of the backpropagation algorithmic computational technique trace back to the seminal works of \citet{linnainmaa1970representation} and \citet{rumelhart1986learning}. Over the past two decades, systems implementing this fundamental algorithm have undergone significant evolution (see Section~\ref{sec:prev-systems}). Given this extensive history, it might seem that the computation of $\nabla f(x)$ has reached its full potential. 
\begin{center}
	\textit{However, our research challenges this assumption, revealing that there are still opportunities for further advancements.}
\end{center}

\subsection{Contributions} 
\label{sec:contributions}


The key contributions of this work are as follows:

\begin{enumerate} 
	
	\item \textbf{Latency-Efficient backpropagation.} \libname{BurTorch} enhances memory access, computational flow of backpropagation to deliver extremely efficient performance in latency-sensitive applications, especially when batch sizes are small or high throughput is difficult to achieve. This makes it well-suited for resource-constrained environments.
	
	\item \textbf{Significant computation gains with a compact implementation.} \libname{BurTorch} delivers substantial speedups and minimizes memory usage, making it ideal for mobile and IoT devices in Federated Learning \citep{FEDLEARN}. These improvements were achieved without increasing implementation complexity, preserving the flexibility for algorithmic refinements and extensions during $\nabla f(x)$ computation.


	
	\item \textbf{Bringing compile-based languages closer to algorithms creation in ML.} High-level scripting languages, while user-friendly, often introduce excessive abstraction in DL systems, limiting system-level optimization across compilers, computing architectures, and operating systems. The runtime of these languages consumes significant system resources \citep{pereira2021ranking}, and the restricted fine-grained control over system resources hampers efforts to minimize memory movement operations. To address these challenges, \libname{BurTorch} eliminates unnecessary layers, reduces reliance on third-party libraries, and optimizes backpropagation performance. Created in modern C++20, \libname{BurTorch} maintains a compact, efficient design while adopting a \libname{PyTorch}-like syntax (see Listing~\ref{lst:exp2-listing} in Appendix~\ref{app:exp2-listing}). Its minimalistic approach avoids the common problem of prolonged compilation times in complex C++ projects. Finally, as large language models evolve, they can assist in generating the necessary system boilerplate code, which could improve the reputation of C++ in research.
	
	\item \textbf{Serialized computations to reduce memory.} Modern frameworks often parallelize batch processing to hide latency, but this increases memory consumption by storing activations for the entire batch, creating a bottleneck during backpropagation \citep{2024fwdllm, mishra2017wrpn} in both image processing and large language model tasks. \libname{BurTorch} addresses this by computing sample-gradient oracles sequentially and overwriting activations across the batch. This reduces the activation memory footprint by a factor proportional to batch size $b$, sacrificing some throughput but minimizing memory usage. In addition to a serialized computation model, \libname{BurTorch} tackles the memory problem by dramatically reducing the code size, which tends to be excessively high in modern DL frameworks.

	\item \textbf{Benchmarking Against Best-Practice Solutions.} We conduct an extensive evaluation of \libname{BurTorch} by benchmarking it against several computational libraries, including \libname{JAX}, \libname{TensorFlow}, \libname{PyTorch}, \libname{Apple MLX}, \libname{Autograd}, \libname{TFLite}, and \libname{Micrograd}. Our experiments span multiple operating systems, utilize various language interfaces, explore different execution modes across these frameworks, and cover a diverse range of tasks. To provide a clear overview of the key improvements, we present a representative summary of the numerical results in Table~\ref{tab:speedup-summary}. As we will see in the main part of the paper, the performance gains of \libname{BurTorch} increase significantly as \( d \) decreases.
	
\end{enumerate}

\begin{table}[h!]
	\footnotesize
	\centering	
	
	\caption{Significant practical performance speedups of \libname{BurTorch} over \libname{PyTorch} on CPU. A representative summary of the expected gains in computing $\nabla f_S(x)$ for small $|S|$, based on the explicit form of $f_S(x)$ in Equation~\eqref{eq:main-est}.}

	\begin{tabular}{|l|c|c|c|c|}
		\hline
		\textbf{Backpropagation Task} & \textbf{Dimension} & \textbf{\makecell[c]{Compute\\ Speedup}} & \textbf{\makecell[c]{Memory\\ Savings}} & \textbf{\makecell[c]{Initialization\\ Speedup}} \\ 
		\hline
		\hline
		1. Small compute graph $f_S(x)$ & \(d \approx 6,000\)   & \(\times 45\)    & \(\times 74\)  & \(\times 354\)  \\ \hline
		2. Medium compute graph $f_S(x)$ & \(d \approx 60,000\)  & \(\times 7.4\)   & \(\times 65\)  & \(\times 340\)  \\ \hline
		3. Large compute graph $f_S(x)$ & \(d \approx 600,000\) & \(\times 1.5\)   & \(\times 37\)  & \(\times 150\)  \\ \hline
		4. Larger compute graph $f_S(x)$ & \(d \approx 1,000,000\) & \(\times 1.2\)   & \(\times 25\)  & \(\times 100\)  \\ \hline
	\end{tabular}
	\label{tab:speedup-summary}
\end{table}

\section{Validation Through Experiments}
\label{sec:experiments}

We present evidence of \libname{BurTorch} achieving low latency and a small memory footprint during $\nabla f(x)$ computation on Windows OS across a series of tasks, starting with the simplest. Results for Linux and macOS are provided in Appendix~\ref{app:extra-experiments-linux} and \ref{app:extra-experiments-macos}, respectively. Additionally, a user-centric energy consumption measurement experiment is included in Appendix~\ref{app:energy-eff}. For detailed software and hardware setups across all experiments, see Appendix~\ref{app:exp-setup}.

\subsection{Tiny compute graph}
\label{sec:experiments-tiny}

\begin{figure}[h!]
	\centering
	\includegraphics[width=1.0\linewidth]{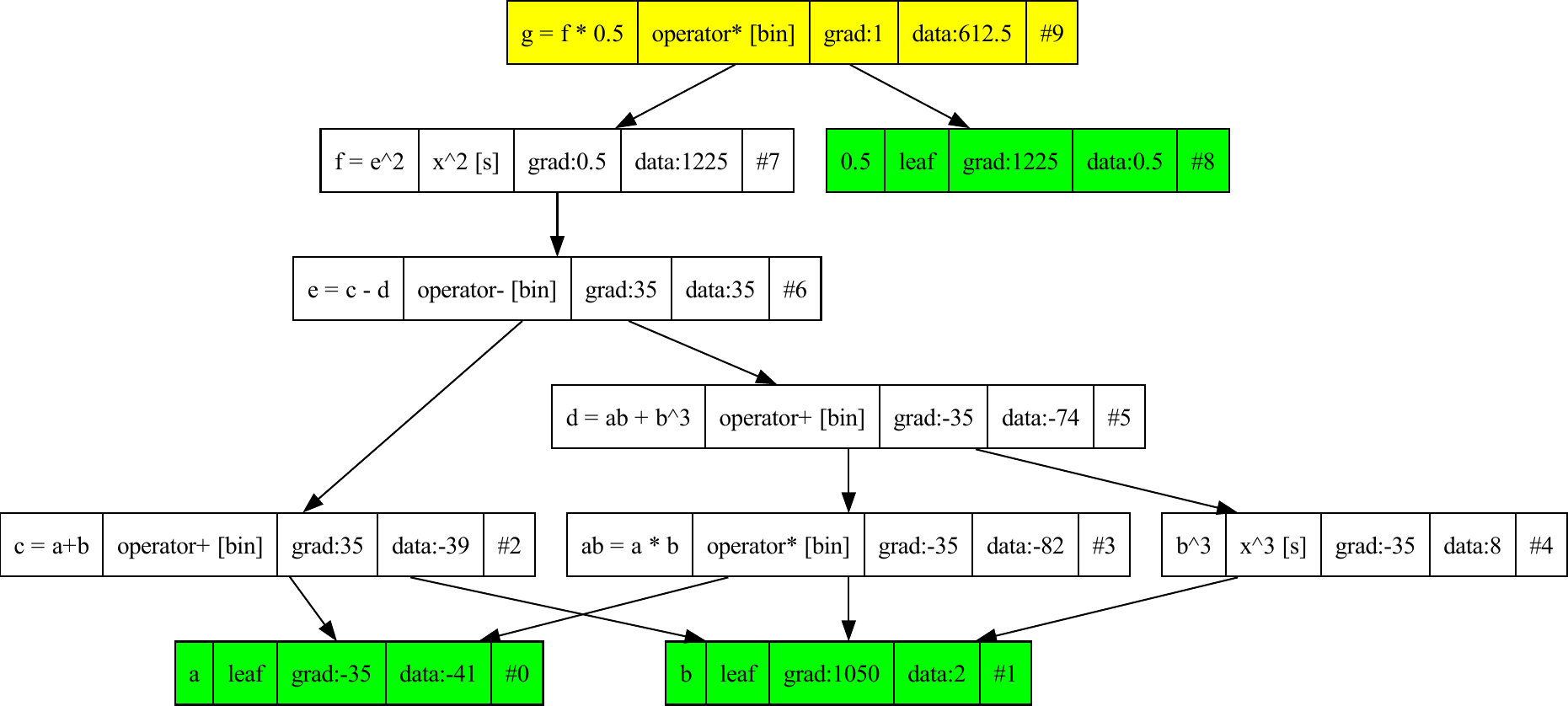}
	\caption{Tiny compute graph with $10$ nodes to evaluate $g=f/2,f=e^2,e=c-d, d=ab + b^3, c=a+b,a=-41,b=2$. Nodes contain: description, operator, $\dfrac{\partial g}{\partial [\mathrm{\textit{node}}]}$, value, raw index. The numerical results across frameworks match exactly.}
	\label{fig:tiny-compute-graph}
\end{figure}

We represent the composite function shown in Figure~\ref{fig:tiny-compute-graph} as an oriented directed tree, where the leaves correspond to constants and variables from $\mathbb{R}$, and the internal nodes represent algebraic operations. When expressing this computation across various frameworks, as detailed in Table~\ref{tab:execution_times_speedup}, we observe a substantial performance gap between \libname{BurTorch} and other state-of-the-art frameworks. \libname{BurTorch} framework consistently outperforms alternatives in computing $\nabla f(x)$ via Automatic Differentiation when handling small graphs.

\begin{figure*}[ht]
\centering
\includegraphics[width=0.85\linewidth]{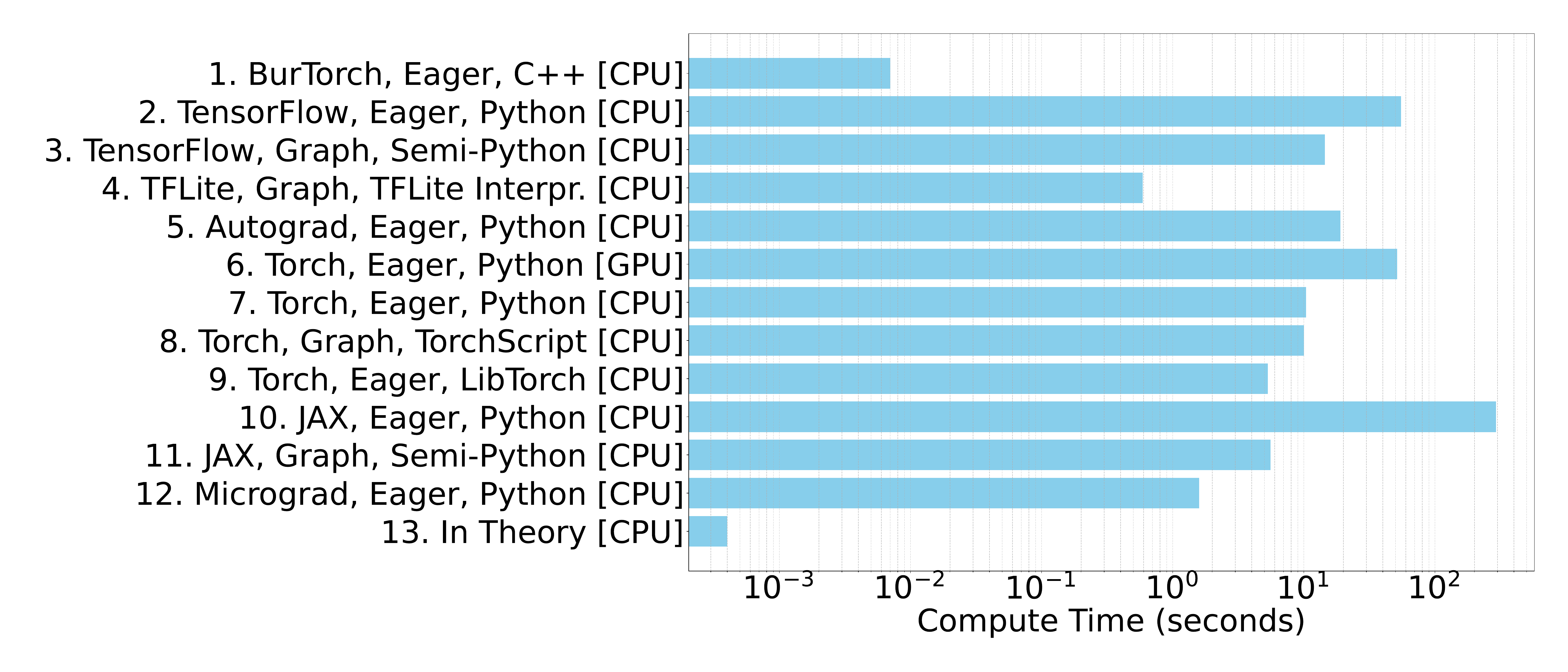}

\caption{Visualization of Table~\ref{tab:execution_times_speedup}. Backpropagation over $100$K iterations with a {tiny} dynamic compute graph from Figure~\ref{fig:tiny-compute-graph}. Computation in FP64, one CPU Core, Windows OS. The numerical results across frameworks match exactly.}

\label{fig:execution_times_speedup}
\end{figure*}

Table~\ref{tab:execution_times_speedup} and Figure~\ref{fig:execution_times_speedup} present backpropagation performance over $10^5$ iterations. The reported time is the mean and standard deviation of five independent runs, with computation performed in FP64 on a single CPU core under Windows OS. The comparison highlights the compute time of different frameworks and languages, along with their relative speedup over \libname{BurTorch}. With the design philosophy outlined in Appendix~\ref{app:design-philosophy} and utilizing research-friendly eager execution, \libname{BurTorch} achieves the fastest time, taking $0.007$ seconds for $10^5$ iterations of backpropagation.

In contrast, other frameworks such as \libname{TensorFlow}, \libname{TF Lite} \citep{abadi2016tensorflow}, \libname{Autograd} \citep{maclaurin2015autograd}, \libname{PyTorch} \citep{paszke2019pytorch}, \libname{JAX} \citep{jax2018github}, and \libname{Micrograd} \citep{karpathy2020micrograd} exhibit significantly higher execution times, with some being up to $7,000$ times slower. The experiment demonstrates that as the compute graph becomes tiny, modern DL frameworks, whether accessed through high-level or optimized APIs, struggle to operate efficiently. The latency associated with transferring control to the GPU only \textit{exacerbates} the issue. These results highlight \libname{BurTorch}'s efficiency in handling tiny-scale $\nabla f(x)$ task on a CPU. Given that a $\times 4$ improvement is considered groundbreaking in Compute Architecture and Compilers, the performance gains of \libname{BurTorch} are significant.

\begin{figure*}
	\centering
	\includegraphics[width=1.0\linewidth]{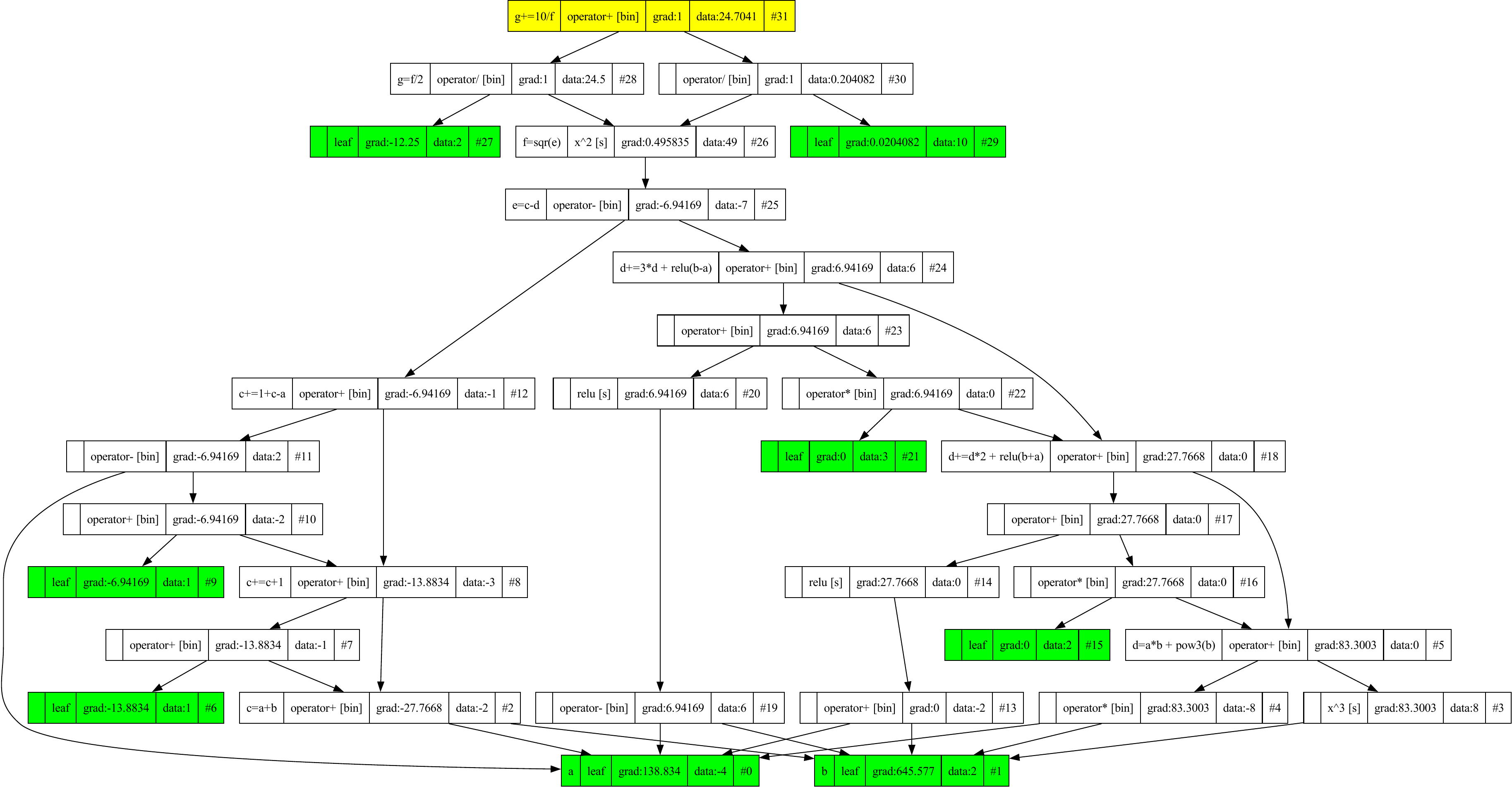}
	\caption{Small compute graph with total $V=32$ nodes and $E=44$ edges to evaluate function from \citet{karpathy2020micrograd}.}

\label{fig:exp2-small-compute-graph}
\end{figure*}

\begin{table*}[h!]
\footnotesize
\centering
\caption{Backpropagation over $100$K iterations with a {tiny} compute graph from Figure~\ref{fig:tiny-compute-graph}. Mean and std. deviation across $5$ launches, FP64, Windows OS. See also Figure~\ref{fig:execution_times_speedup}. The numerical results across frameworks match exactly.
}
\begin{tabular}{|l|l|l|l|l|}
	\hline
	\textbf{\#} & \textbf{Framework, Mode, Language} & \textbf{Device} & \textbf{\makecell[c]{Compute Time \\ (sec.)}} & \textbf{\makecell[c]{Relative \\ to \\ BurTorch}} \\ 
	\hline
	\hline
	\cellcolor{bgcolorwe}1&\cellcolor{bgcolorwe}BurTorch, Eager, C++ & \cellcolor{bgcolorwe}CPU                   & \cellcolor{bgcolorwe}$0.007 \pm 0.0004$             & \cellcolor{bgcolorwe}$\times 1.0$ (We)             \\ \hline
	2&TensorFlow 2.8.0, Eager, Python & CPU          & $55.217 \pm 0.2975$         & $\times 7\,888.1\,\,\,$                    \\ \hline
	3&TensorFlow 2.8.0, Graph, Semi-Python & CPU          & $14.469 \pm 0.0734$         & $\times 2\,067.0\,\,\,$                    \\ \hline
	4&\makecell[l]{TF Lite 2.8.0, Graph, TF Lite Interpreter} & CPU          & $0.589 \pm 0.0102$         & $\times 84\,\,\,$                    \\ \hline
	5&Autograd 1.7.0, Eager, Python & CPU            & $18.956 \pm 0.2962$        & $\times 2\,708.0\,\,\,$                    \\ \hline
	6&{PyTorch} 2.5.1, Eager, Python & \textbf{GPU}          & $51.380 \pm 0.4666$              & $\times 7\,340.0\,\,\,$                    \\ \hline
	7&{PyTorch} 2.5.1, Eager, Python & CPU          & $10.419 \pm 0.0647$              & $\times 1\,488.4\,\,\,$                    \\ \hline
	8&{PyTorch} 2.5.1, Graph, TorchScript & CPU & $9.994 \pm 0.1021$                 & $\times 1\,428.5\,\,\,$                    \\ \hline
	9&{PyTorch} 2.5.1, Eager, LibTorch, C++ & CPU & $5.300 \pm 0.0667$                 & $\times 757.14\,\,\,$                    \\ \hline
	10&{JAX} 0.4.30, Eager, Python & CPU                & $291.764 \pm 8.5373$         & $\times 41\,860.5$                      \\ \hline
	11&{JAX} 0.4.30, Graph, Semi-Python & CPU                & $5.580 \pm 0.0661$         & $\times 797.1\,\,\,\,\,\,\,$                      \\ \hline
	12&Micrograd, Eager, Python & CPU                & $1.590 \pm 0.0152$                & $\times 227.1\,\,\,\,\,\,\,$                      \\ \hline
	\cellcolor{bgcolorwe}13&\cellcolor{bgcolorwe} In Theory for this CPU (Registers Only) & \cellcolor{bgcolorwe}CPU & \cellcolor{bgcolorwe}$\Omega(0.0004)$             & \cellcolor{bgcolorwe}$\times 0.057$ \\ \hline
\end{tabular}
\label{tab:execution_times_speedup}
\end{table*}

\begin{table*}[h!]
\footnotesize
\centering
\caption{Backpropagation over $20$K iterations with a \textit{small} dynamically constructed compute graph (Figure~\ref{fig:exp2-small-compute-graph}) with $32$ nodes, $5$ trials, FP64, Windows OS. The numerical results across frameworks match exactly.}

\begin{tabular}{|c|l|l|l|l|l|}
	\hline
	\textbf{\parbox{1.7cm}{\center \footnotesize{Framework,\\Mode,\\Device}}} &
	\textbf{\parbox{1.5cm}{\center \footnotesize{Compute Time\\(sec.)}}} & 
	\textbf{\parbox{1.7cm}{\center \footnotesize{Minimum Compute Time\\(sec.)}}} & 
	\textbf{\parbox{1.7cm}{\center \footnotesize{Total\\CPU Clocks \\ ($10^6$ ticks)}}} & 
	\multicolumn{2}{c|}{\makecell[c]{\textbf{Peak Memory}\\\textbf{(MB)}}} \\ 
	\cline{5-6}
	& & & & \textbf{Private} & \textbf{Resident} \\ 
	\hline
	\hline
	
	{\cellcolor{bgcolorwe} \makecell[c]{\cellcolor{bgcolorwe} BurTorch,\,\,\,\,\,\,\,\,\,\,\\ Eager, CPU}}&
	
	{\cellcolor{bgcolorwe}\begin{tabular}{@{}l@{}} \cellcolor{bgcolorwe} $0.0082$ \\ \cellcolor{bgcolorwe} $\pm 0.0003$ \end{tabular}} &
	\cellcolor{bgcolorwe}\begin{tabular}{@{}l@{}} {\color{abscolor} abs: $0.008$} \\ rel: $\times 1$ (We) \end{tabular} & 
	\cellcolor{bgcolorwe}\begin{tabular}{@{}l@{}} {\color{abscolor} abs: $39$} \\ rel: $\times 1$ \end{tabular} & 
	\cellcolor{bgcolorwe}\begin{tabular}{@{}l@{}} {\color{abscolor} abs: $0.6$} \\ rel: $\times 1$ \end{tabular} & \cellcolor{bgcolorwe}\begin{tabular}{@{}l@{}} {\color{abscolor} abs: $3.9$} \\ rel: $\times 1$ \end{tabular} \\ 
	\hline
	
	\makecell[c]{TensorFlow,\\Eager, CPU}&
	\begin{tabular}{@{}l@{}} $24.763$ \\ $\pm 0.5533$ \end{tabular} &
	\begin{tabular}{@{}l@{}} {\color{abscolor} abs: $24.158$} \\ rel: $\times 3019.7$ \end{tabular} & 
	\begin{tabular}{@{}l@{}} {\color{abscolor} abs: $75\,207$} \\ rel: $\times 1928.3$ \end{tabular} & 
	\begin{tabular}{@{}l@{}} {\color{abscolor} abs: $2120.8$} \\ rel: $\times 3534.6$ \end{tabular} & \begin{tabular}{@{}l@{}} {\color{abscolor} abs: $497.2$} \\ rel: $\times 127.5$ \end{tabular} \\ 
	\hline
	
	\makecell[c]{TensorFlow,\\Graph, CPU}&
	\begin{tabular}{@{}l@{}} $4.772$ \\ $\pm 0.4495$ \end{tabular} &
	\begin{tabular}{@{}l@{}} {\color{abscolor} abs: $3.874$} \\ rel: $\times 484.2$ \end{tabular} & 
	\begin{tabular}{@{}l@{}} {\color{abscolor} abs: $18\,065$} \\ rel: $\times 463.2$ \end{tabular} & 
	\begin{tabular}{@{}l@{}} {\color{abscolor} abs: $2121.6$} \\ rel: $\times 3536.0$ \end{tabular} & \begin{tabular}{@{}l@{}} {\color{abscolor} abs: $603.8$} \\ rel: $\times 154.82$ \end{tabular} \\ 
	\hline
	
	\makecell[c]{TF Lite,\\Graph, CPU}&
	\begin{tabular}{@{}l@{}} $2.435$ \\ $\pm 0.022$ \end{tabular} &
	\begin{tabular}{@{}l@{}} {\color{abscolor} abs: $2.400$} \\ rel: $\times 300.0$ \end{tabular} & 
	\begin{tabular}{@{}l@{}} {\color{abscolor} abs: $13\,624$} \\ rel: $\times 349.3$ \end{tabular} & 
	\begin{tabular}{@{}l@{}} {\color{abscolor} abs: $2247.1$} \\ rel: $\times 3745.1$ \end{tabular} & \begin{tabular}{@{}l@{}} {\color{abscolor} abs: $583.7$} \\ rel: $\times 149.6$ \end{tabular} \\ 
	\hline
	
	\makecell[c]{Autograd,\\Eager, CPU}&
	\begin{tabular}{@{}l@{}} $12.100$ \\ $\pm 0.0855$ \end{tabular} &
	\begin{tabular}{@{}l@{}} {\color{abscolor} abs: $11.971$} \\ rel: $\times 1496.3$ \end{tabular} & 
	\begin{tabular}{@{}l@{}} {\color{abscolor} abs: $36\,817$} \\ rel: $\times 944.0$ \end{tabular} & 
	\begin{tabular}{@{}l@{}} {\color{abscolor}  abs: $708.7$} \\ rel: $\times 1181.1$ \end{tabular} & \begin{tabular}{@{}l@{}} {\color{abscolor} abs: $29.7$} \\ rel: $\times 7.6$ \end{tabular} \\ 
	\hline
	
	\makecell[c]{PyTorch,\\Eager, \textbf{GPU}} &
	\begin{tabular}{@{}l@{}} $34.932$ \\ $\pm 0.1613$ \end{tabular} &
	\begin{tabular}{@{}l@{}} {\color{abscolor} abs: $34.784$} \\ rel: $\times 4348.0$ \end{tabular} & 
	\begin{tabular}{@{}l@{}} {\color{abscolor} abs: $83\,501$} \\ rel: $\times 2141.0$ \end{tabular} & 
	\begin{tabular}{@{}l@{}} {\color{abscolor} abs: $1639.5$} \\ rel: $\times 2732.5$ \end{tabular} & \begin{tabular}{@{}l@{}} {\color{abscolor} abs: $503.7$} \\ rel: $\times 129.15$ \end{tabular} \\ 
	\hline
	
	\makecell[c]{PyTorch,\\Eager, CPU,\\Python}&
	\begin{tabular}{@{}l@{}} $5.438$ \\ $\pm 0.0191$ \end{tabular} &
	\begin{tabular}{@{}l@{}} {\color{abscolor} abs: $5.418$} \\ rel: $\times 677.3$ \end{tabular} & 
	\begin{tabular}{@{}l@{}} {\color{abscolor} abs: $18\,977$} \\ rel: $\times 486.6$ \end{tabular} & 
	\begin{tabular}{@{}l@{}} {\color{abscolor}  abs: $1235.5$} \\ rel: $\times 2059.2$ \end{tabular} & \begin{tabular}{@{}l@{}} {\color{abscolor} abs: $334.3$} \\ rel: $\times 85.71$ \end{tabular} \\ 
	\hline
	
	\makecell[c]{PyTorch,\\ Graph, CPU, \\ TorchScript} &
	\begin{tabular}{@{}l@{}} $5.529$ \\ $\pm 0.0244$ \end{tabular} &
	\begin{tabular}{@{}l@{}} {\color{abscolor} abs: $5.505$} \\ rel: $\times 688.1$ \end{tabular} & 
	\begin{tabular}{@{}l@{}} {\color{abscolor} abs: $18\,956$} \\ rel: $\times 486.0$ \end{tabular} & 
	\begin{tabular}{@{}l@{}} {\color{abscolor} abs: $1245.7$} \\ rel: $\times 2076.2$ \end{tabular} &  \begin{tabular}{@{}l@{}} {\color{abscolor} abs: $346.8$} \\ rel: $\times 88.92$ \end{tabular} \\ 
	\hline
	
	{\centering {\begin{tabular}{@{}c@{}} PyTorch,\\Eager,CPU\\ LibTorch \end{tabular}}} &
	\begin{tabular}{@{}l@{}} $3.203$ \\ $\pm 0.0351$ \end{tabular} &
	\begin{tabular}{@{}l@{}} {\color{abscolor} abs: $3.155$} \\ rel: $\times 394.4$ \end{tabular} & 
	\begin{tabular}{@{}l@{}} {\color{abscolor} abs: $10\,283$} \\ rel: $\times 263.7$ \end{tabular} & 
	\begin{tabular}{@{}l@{}} {\color{abscolor} abs: $74.6$} \\ rel: $\times 124.3$ \end{tabular} & \begin{tabular}{@{}l@{}} {\color{abscolor} abs: $77.5$} \\ rel: $\times 19.9$ \end{tabular} \\ 
	\hline
	
	\makecell[c]{JAX,\\Eager, CPU} &
	\begin{tabular}{@{}l@{}} $185.862$ \\ $\pm 1.8866$ \end{tabular} &
	\begin{tabular}{@{}l@{}} {\color{abscolor} abs: $182.796$} \\ rel: $\times 22849.5$ \end{tabular} & 
	\begin{tabular}{@{}l@{}} {\color{abscolor} abs: $597\,088$} \\ rel: $\times 15\,309.9$ \end{tabular} & 
	\begin{tabular}{@{}l@{}} {\color{abscolor} abs: $791.3$} \\ rel: $\times 1318.8$ \end{tabular} & \begin{tabular}{@{}l@{}} {\color{abscolor} abs: $134.8$} \\ rel: $\times 34.5$ \end{tabular} \\ 
	\hline
	
	\makecell[c]{JAX,\\Graph, CPU} &
	\begin{tabular}{@{}l@{}} $1.184$ \\ $\pm 0.0189$ \end{tabular} &
	\begin{tabular}{@{}l@{}} {\color{abscolor} abs: $1.159$} \\ rel: $\times 144.9$ \end{tabular} & 
	\begin{tabular}{@{}l@{}} {\color{abscolor} abs: $4\,544$} \\ rel: $\times 116.5$ \end{tabular} & 
	\begin{tabular}{@{}l@{}} {\color{abscolor} abs: $765.2$} \\ rel: $\times 1275.3$ \end{tabular} & \begin{tabular}{@{}l@{}} {\color{abscolor} abs: $111.2$} \\ rel: $\times 28.5$ \end{tabular} \\ 
	\hline
	
	\makecell[c]{Micrograd,\\Eager, CPU}&
	\begin{tabular}{@{}l@{}} $1.071$ \\ $\pm 0.0074$ \end{tabular} &
	\begin{tabular}{@{}l@{}} {\color{abscolor} abs: $1.062$} \\ rel: $\times 132.8$ \end{tabular} & 
	\begin{tabular}{@{}l@{}} {\color{abscolor} abs: $3\,193$} \\ rel: $\times 81.9$ \end{tabular} & 
	\begin{tabular}{@{}l@{}} {\color{abscolor} abs: $6.2$} \\ rel: $\times 10.3$ \end{tabular} & \begin{tabular}{@{}l@{}} {\color{abscolor} abs: $12.2$} \\ rel: $\times 3.1$
	\end{tabular} \\
	\hline
	\cellcolor{bgcolorwe}\makecell[c]{In Theory, \\ Registers only,\,\\this CPU} & \multicolumn{2}{|c|}{\cellcolor{bgcolorwe} $\Omega(0.00033)$} &\cellcolor{bgcolorwe}{\color{abscolor}abs: $1.52$}   &
	\multicolumn{2}{|c|}{\cellcolor{bgcolorwe} \color{abscolor}abs: $0$}
	\\ \hline
	
	\hline
\end{tabular}
\label{tab:exp2-execution-times-speedup}
\end{table*}

\subsection{Small compute graph} 
\label{sec:exp2-small-compute-graphs}

In this experiment, we evaluate the performance of various frameworks on a small computation graph in Figure~\ref{fig:exp2-small-compute-graph} with 32 nodes, adapted from \citet{karpathy2020micrograd}. The comparison includes \libname{BurTorch}, \libname{TensorFlow}, \libname{TF Lite}, \libname{Autograd}, \libname{PyTorch}, \libname{Micrograd}, and \libname{JAX} running on Windows OS. The results are summarized in Table~\ref{tab:exp2-execution-times-speedup}. While popular frameworks, when utilized through C++ or Python (and Python-like languages via built-in compilers), exhibit nearly identical performance, \libname{BurTorch} stands out due to its faster computation. 

\libname{BurTorch} significantly reduces various forms of end-to-end memory allocation from the OS. We observe this in two key memory metrics. The first is the \textit{peak virtual private committed memory}. This represents the memory reserved by the OS, either in DRAM or in the swap file. While it may not be used by the application, Windows OS cannot reclaim it. This memory is non-shareable and cannot be reused by other applications. The second metric is the \textit{resident memory} (also referred to as the \textit{working set}), which is the memory allocated by the OS for applications from installed DRAM. Switching from \libname{Python} to \libname{TorchScript} and C++ with \libname{PyTorch} does lead to performance improvements, but the gains are not dramatic. This suggests that the issue with \libname{PyTorch} in this context extends beyond the language choice.


\subsection{Small compute graphs: loading and saving}
\label{sec:exp3-small-compute-graphs}

\begin{table}[h!]
\footnotesize
\small
\centering
\caption{Save and load subset of $7$ compute graph activations over $5 \cdot 10^3$ iterations to disk. The experiment uses a \textit{small}, dynamically constructed compute graph (Figure~\ref{fig:exp2-small-compute-graph}). Activations and gradients are computed in FP64 format. Raw payload size: $56$ bytes, $5$ trials.}
\begin{tabular}{|l|l|l|l|l|l|}
	\hline
	\textbf{\#} &
	\textbf{{\footnotesize{Framework, Mode, Language}}} & 
	\textbf{{\footnotesize{\parbox{1.5cm}{Save\\ (sec.)}}}} & 
	\textbf{{\footnotesize{\parbox{1.5cm}{Load\\ (sec.)}}}} & 
	\textbf{{\footnotesize\parbox{1.5cm}{{File Size\\ (bytes)}}}}
	\\ 
	\hline
	\hline
	
	\cellcolor{bgcolorwe}1 & \cellcolor{bgcolorwe}BurTorch, Eager, C++ 
	&\cellcolor{bgcolorwe}0.75 
	&\cellcolor{bgcolorwe}0.08
	&\cellcolor{bgcolorwe}56
	\\ 
	\hline

	2 & TensorFlow, Eager, Python 
	& 1.97
	& 1.66
	& 634
	\\ 
	\hline

	3 & \begin{tabular}{@{}l@{}} 
		TensorFlow, Graph, Semi-Python 
	\end{tabular}
	& 1.20
	& 0.30
	& 329
	\\ 
	\hline
	
	4 & \begin{tabular}{@{}l@{}} PyTorch, Eager, Python \\ {PyTorch} $\to$ {NumPy} $\to$ Bytes \end{tabular}
	& 1.11
	& 0.22
	& 56
	\\ 
	\hline

	5 & PyTorch, Eager, Python 
	& 2.54
	& 1.36
	& 2564
	\\ 
	\hline
	
	6 & PyTorch, Graph, TorchScript
	& 2.52
	& 1.34
	& 2564
	\\ 
	\hline
	
	7 & PyTorch, Eager, LibTorch, C++ 
	& 1.55
	& 0.95
	& 3569
	\\ 
	\hline

	\hline
\end{tabular}
\label{tab:exp2-save-load-speedup}
\end{table}

In this experiment, we assess the performance of several frameworks in terms of saving and loading activations (both leaf and intermediate nodes) from a small, dynamically constructed compute graph in Figure~\ref{fig:exp2-small-compute-graph} with 32 nodes from Section \ref{sec:exp2-small-compute-graphs}. The results are presented in Table~\ref{tab:exp2-save-load-speedup}.

The compute graph is used to simulate the saving and loading of activations over $5000$ iterations, and we have compared the performance of \libname{BurTorch} against \libname{TensorFlow}, \libname{PyTorch}, and \libname{JAX}. This experiment is conducted in the same hardware and software environment as the previous one. Activations from seven nodes, labeled (a)–(g) in Figure~\ref{fig:exp2-small-compute-graph} are saved. This results in a raw payload size of 56 bytes. Frameworks like \libname{Micrograd}, \libname{Autograd}, and \libname{JAX} are excluded from the comparison due to their lack of native support for saving computation graphs. \libname{BurTorch} not only significantly reduces save and load times but also minimizes the size of data transferred to and from disk.

\subsection{{BurTorch} on medium compute graphs}
\label{sec:exp3-medium-compute-graphs}

We evaluated \libname{BurTorch} on a character-level autoregressive prediction model, designed as a medium-complexity compute graph, based on the architecture from \citet{bengio2000neural}. This model uses English characters, each represented as a trainable embedding variable from $\mathbb{R}^{64}$. The dataset consists of 26 characters and a special token for sequence start, end, and padding, totaling 27 tokens. The context length (block size) for predicting the next-token is set to $16$. All trainable parameters, activations, and computed derivatives are represented using FP32. The training dataset, from \citet{makemore2023}, contained $n=228,146$ samples.

We compared \libname{BurTorch} with a \libname{PyTorch} implementation from \citet{makemore2023}. Tables \ref{tab:exp3-compute-and-mem-speedup-b1} and \ref{tab:exp3-compute-and-mem-speedup-b64} show the performance comparison for models with varying numbers of trainable parameters, ranging from $5,963$ to $1,079,003$.

During the experiments, we train using \algname{SGD} with the gradient estimator given by Equation~\ref{eq:main-grad-est}:
	
$$x^{k+1}=x^{k} - \gamma \dfrac{1}{b}\sum_{i\in S_k} \nabla f_i(x^k),$$
where $S_k \sim_{\mathrm{u.a.r}} \{s:s \in 2^{[n]} \land |s|=b\}$. Tables \ref{tab:exp3-compute-and-mem-speedup-b1} and \ref{tab:exp3-compute-and-mem-speedup-b64} report the results for batch sizes of $b=1$ and $b=64$, respectively. Trainable pairs for the autoregressive model were sampled uniformly at random from the training dataset. The model size was adjusted by modifying the number of hidden units in the first layer and the corresponding input units in the second layer (both layers use tanh activation).

\paragraph{Initialization time.} This time refers to the total time required for setting up the training process, including data loading, preprocessing, one backpropagation iteration (considering \libname{PyTorch’s} deferred memory allocations), and deinitialization. As shown in Table~\ref{tab:exp3-compute-and-mem-speedup-b1}, \libname{PyTorch's} initialization time for a model with $5,963$ parameters was equivalent to $3,794$ gradient oracle computations with $b=1$, whereas \libname{BurTorch} reduced this time to $488$ gradient oracles.

\paragraph{Memory efficiency.} We measured peak private virtual memory usage, representing the maximum memory allocated exclusively for the training process. For a compute graph with $1,079,003$ parameters, \libname{BurTorch} required $107$ MB of memory only, while \libname{PyTorch} required over $2.7$ GB for both $b=1$ and $b=64$. This highlights that the minimalist design of \libname{BurTorch} significantly reduces memory consumption, making it suitable for resource-constrained environments.

\paragraph{Compute time.} The reported time is the estimated mean and standard deviation over $4000$ iterations of \algname{SGD} as a computational model. \libname{BurTorch} consistently outperformed \libname{PyTorch} in terms of expected compute time for compute graphs with $d \leq 10^6$ parameters. For example, for $d=5,963$ and $b=1$, \libname{PyTorch} was approximately $45$ times slower than \libname{BurTorch}. The complex architecture of \libname{PyTorch} led to a more variable standard deviation in execution time.

\paragraph{Comparative analysis of computation time.} 

As shown in Table~\ref{tab:exp3-compute-and-mem-speedup-b1}, \libname{BurTorch} consistently outperforms \libname{PyTorch} for $b=1$ with $d$ ranging from 6K to 1M. While \libname{BurTorch} demonstrates a clear performance advantage for smaller compute graphs or small batch sizes, the gap narrows as the model size exceeds 100K parameters for $b=64$. This is because \libname{PyTorch} implementation leverages techniques such as optimized kernels and vectorized operations designed for maximum throughput.

\paragraph{Comparative analysis of memory usage.} Limited memory in edge devices poses challenges during inference \citep{laskaridis2024melting} and becomes even more critical during training. If peak memory usage exceeds available capacity and $\nabla f(x)$ computation cannot be controlled, it may \textit{seriously} disrupt the execution.

\begin{table*}[h!]
\footnotesize
\centering
\caption{Comparison of \libname{BurTorch} and \libname{PyTorch} performance for training MLP-like model. Batch: $b=1$, Compute: FP32, Single CPU core. Initialization time is end-to-end time for training with $1$ iteration. Compute time excludes batch preparation. Memory is the peak private virtual memory.}
\label{tab:exp3-compute-and-mem-speedup-b1}
\begin{tabular}{|l|l|l|l|l|l|l|l|}
	\hline
	\textbf{\#} &\textbf{Parameters (d)} & \multicolumn{3}{c|}{\textbf{\makecell[c]{PyTorch,\\ Eager, v2.5.1 [CPU]}}} & \multicolumn{3}{c|}{\cellcolor{bgcolorwe}{\textbf{{BurTorch, Eager [CPU]}}}} \\ 
	\cline{3-8}
	&\textbf{Hidden Dim.(e)} & \textbf{\makecell[c]{Init\\(ms)}} & \textbf{\makecell[c]{Compute\\(ms)}} & \textbf{\makecell[c]{Mem.\\(MB)}} & {\textbf{\makecell[c]{Init\\(ms)}}} & {\textbf{\makecell[c]{Compute\,\,\,\,\,\,\\(ms)}}} & {\textbf{\makecell[c]{Mem.\\(MB)}}} \\ 
	\hline
	\hline
	1&$5,963\ (e=4)$ & $5\,540$ & $1.46 \pm 4.63$ & $2\,651$ & $15.63$ & $0.032 \pm 0.008$ & $35.8$ \\ 
	2&$18,587\ (e=16)$ & $5\,627$ & $1.52 \pm 4.21$ & $2\,653$ & $16.51$ & $0.074 \pm 0.016$ & $36.7$ \\ 
	3&$35,419\ (e=32)$ & $5\,673$ & $1.55 \pm 5.00$ & $2\,653$ & $18.24$ & $0.124 \pm 0.019$ & $38.3$ \\ 
	4&$69,083\ (e=64)$ & $5\,537$ & $1.63 \pm 4.62$ & $2\,668$ & $18.94$ & $0.221 \pm 0.040$ & $40.8$ \\ 
	5&$136,411\ (e=128)$ & $5\,799$ & $1.79 \pm 5.19$ & $2\,660$ & $21.39$ & $0.417 \pm 0.077$ & $45.9$ \\ 
	6&$540,379\ (e=512)$ & $5\,556$ & $3.01 \pm 5.57$ & $2\,683$ & $37.09$ & $2.093 \pm 0.429$ & $71.4$ \\ 
	7&$1,079,003\ (e=1024)$ & $5\,544$ & $5.57 \pm 6.75$ & $2\,719$ & $56.57$ & $4.550 \pm 0.847$ & $107.0$ \\ 
	\hline
\end{tabular}
\end{table*}

\begin{table*}[h!]
\footnotesize
\centering
\caption{Comparison of \libname{BurTorch} and \libname{PyTorch} performance for training MLP-like model. Batch: $b=64$, Compute: FP32, Single CPU core. Initialization time is end-to-end time for training with $1$ iteration. Compute time excludes batch preparation. Memory is the peak private virtual memory.}
\label{tab:exp3-compute-and-mem-speedup-b64}
\begin{tabular}{|l|l|l|l|l|l|l|l|}
	\hline
	\textbf{\#} & \textbf{Parameters (d)} & \multicolumn{3}{c|}{\textbf{\makecell[c]{PyTorch,\\ Eager, v2.5.1 [CPU]}}} & \multicolumn{3}{c|}{\cellcolor{bgcolorwe}{\textbf{{BurTorch, Eager [CPU]}}}} \\ 
	\cline{3-8}
	&\textbf{Hidden Dim.(e)} & \textbf{\makecell[c]{Init\\(ms)}} & \textbf{\makecell[c]{Compute\\(ms)}} & \textbf{\makecell[c]{Mem.\\(MB)}} & {\textbf{\makecell[c]{Init\\(ms)}}} & {\textbf{\makecell[c]{Compute\\(ms)}}} & {\textbf{\makecell[c]{Mem.\\(MB)}}} \\ 
	\hline
	\hline
	1&$5,963\ (e=4)$ & $5\,658$ & $6.17 \pm 4.71$ & $2\,678$ & $17.33$ & $0.52 \pm 0.05$ & $35.8$ \\ 
	2&$18,587\ (e=16)$ & $5\,774$ & $6.39 \pm 4.57$ & $2\,678$ & $17.38$ & $1.70 \pm 0.12$ & $36.7$ \\ 
	3&$35,419\ (e=32)$ & $5\,941$ & $7.79 \pm 4.40$ & $2\,679$ & $18.29$ & $3.25 \pm 0.17$ & $38.3$ \\ 
	4&$69,083\ (e=64)$ & $6\,020$ & $9.43 \pm 4.81$ & $2\,686$ & $23.62$ & $6.50 \pm 0.31$ & $40.8$ \\ 
	5&$136,411\ (e=128)$ & $6\,176$ & $13.01 \pm 4.88$ & $2\,695$ & $35.62$ & $13.47 \pm 0.71$ & $45.4$ \\ 
	6&$540,379\ (e=512)$ & $6\,151$ & $34.97 \pm 4.05$ & $2\,698$ & $102.19$ & $64.17 \pm 3.22$ & $71.4$ \\ 
	7&$1,079,003\ (e=1024)$ & $5\,926$ & $64.61 \pm 6.10$ & $2\,723$ & $197.67$ & $145.87 \pm 6.04$ & $106.9$ \\ 
	\hline
\end{tabular}
\end{table*}


\subsection{{BurTorch} on a GPT-3-like computation graph} 
\label{sec:exp4-burtorch-fot-gpt3}

Recent advances in Large Language Models (LLMs) indicate significant progress toward artificial general intelligence \citep{bubeck2023sparks}. These models enable intuitive natural language interactions. Their training relies on large-scale data collection, careful tokenization strategy, and fine-tuning for question-answer alignment. First-order continuous optimization, implemented through gradient estimates as given in Equation~\ref{eq:main-grad-est}, is fundamental to their training and fine-tuning.

\paragraph{\color{black}{Experiment setup for GPT-3-like model.}}
\libname{BurTorch} offers the essential computational primitives to support gradient oracle computation while training Transformer-based models. We demonstrate this by constructing a \modelname{GPT}-like model, following the architectures of \citet{transformer} and \modelname{GPT-3} \citep{gpt}. This decoder-only Transformer is designed for the next-token character generation, with parameter counts ranging from $125 \cdot 10^6$ to $125 \cdot 10^9$ \citep{gpt}. To ensure computational feasibility, we significantly scaled down the original \modelname{GPT-3} model by modifying the configuration parameters outlined in Table 2.1 of \citet{gpt}. Although the scalar-level computation steps in the \modelname{GPT-3} architecture are complex, they can be expressed using the primitives detailed in Appendix~\ref{app:burotrch-scalar-operators} and \ref{app:burotrch-derived-operators}. For additional information on Transformer architectures, see \citet{transformer, scardapane2024alice, bishop2023deep}.

\paragraph{Input.} In this experiment, input characters from the Shakespeare dataset \citep{tinyshakespeare} are tokenized into integers, consisting of 65 unique ASCII characters (including padding), resulting in a vocabulary size of $V = 65$. The input sequence (context length) tokens are processed, and the corresponding output sequence (context length) tokens are generated. Once a specific sentence is sampled, the corresponding trainable token embeddings and positional embeddings are added elementwise at each position in the context, without additional transformations.

\paragraph{Internals.}
Next, the sequence of characters with a fixed length, known as the block size or context length, is processed. This represents a typical sequence of tokens handled by the Transformer architecture.

Internally, the generative model consists of multiple layers, which are named Transformer encoder blocks. In our experiment, we use six, but in general, it is a meta-parameter. Each Transformer encoder \citep{transformer} performs the following computations:
(a) six self-attention heads, followed by a concatenation of their outputs and an affine transformation into the embedding space.
(b) two residual connections.
(c) two layer normalization layers.
(d) one two-layer feed-forward neural network with affine transformations.

\paragraph{Output.} 
The \modelname{GPT-3} text generation process involves passing the input context through a series of encoder blocks. After this, finally, each token in the context length block is processed with an affine transformation to generate logits in a vector space of size $V$.

Logits in this space are processed with a softmax function to produce the probability distribution for the token. The loss function, $CE(p,\hat{p}) = -\sum_{i=1}^{V} p_i \log(\hat{p_i})$, compares the true probability mass function (p.m.f.) for a next-token in sequence and predicted p.m.f.

\paragraph{GPT-3-like model: configuration.}

The miniaturized \modelname{GPT-3} configuration which has been used in experiments includes: (i) \texttt{n\_layer=6} Multi-head Self-Attention encoder blocks; (ii) \texttt{k\_heads=6} heads in each block; (iii) \texttt{k\_block\_size=8} for context length (input sequence length); (iv) \texttt{d\_model=24} for the embedding dimension. Computation was performed in FP32 on a single-core CPU, with a total of $46,289$ trainable parameters. The number of \algname{SGD} iterations is $3000$.

\paragraph{GPT-3-like model: experimental results.}

The results in Table~\ref{tab:exp4-compute-and-mem-speedup-win} compare \libname{BurTorch} with a baseline \libname{Python} implementation \citet{karpathygptimplementation} of a \modelname{GPT-3}-type model, focusing on peak memory consumption and the mean and standard deviation of the computation time for a single \algname{SGD} oracle with a fixed batch size over $3$K \algname{SGD} oracles. Additionally, we evaluate its performance against \libname{PyTorch} implementation optimized with \libname{TorchScript}, a built-in compilation technique in \libname{PyTorch} (for more details see Appendix~\ref{app:torch-compile-techniques-backprop-speed-cpu}).

As shown in Table~\ref{tab:exp4-compute-and-mem-speedup-win}, \libname{BurTorch} achieves a $\times 20$ speedup with a batch size of $1$ and reduces memory requirements by $\times 100$. The memory metric represents peak virtual private memory usage. Notably, \libname{BurTorch} also exhibits lower execution time variance than the baseline. By processing samples sequentially, it avoids storing all activations simultaneously, preventing memory usage from scaling with batch size. As batch size increases to $64$, \libname{PyTorch} design outperforms \libname{BurTorch} in terms of time per batch by $\times 1.4$.

\begin{table*}
\footnotesize
\centering
\caption{\libname{BurTorch} and \libname{PyTorch} in training \modelname{GPT-3} like model, FP32, 1 CPU core, Peak private virtual memory. Trainable variables: $46$K.}
\label{tab:exp4-compute-and-mem-speedup-win}
\begin{tabular}{|l|c|c|c|c|c|c|}
	\hline
	\textbf{Batch} & \multicolumn{2}{c|}{\cellcolor{bgcolorwe}{\textbf{BurTorch, Eager, C++}}} & \multicolumn{2}{c|}{\textbf{\makecell[c]{PyTorch,\\ Graph, TorchScript}}} & \multicolumn{2}{c|}{\textbf{\makecell[c]{PyTorch,\\ Eager, Python}}} \\ 
	\cline{2-7}
	& {\textbf{\makecell[c]{Compute\\(ms)}}} & {\textbf{\makecell[c]{Mem.\\(MB)}}} & \textbf{\makecell[c]{Compute\\(ms)}} & \textbf{\makecell[c]{Mem.\\(MB)}} & \textbf{\makecell[c]{Compute\\(ms)}} & \textbf{\makecell[c]{Mem.\\(MB)}} \\ 
	\hline
	\hline
	$1$ & $0.515 \pm 0.067$ & $16.7$ & $11.119 \pm 48.118$ & $1\,624$ & $11.715 \pm 10.741$ & $1\,300$ \\ 
	$2$ & $1.027 \pm 0.091$ & $16.7$ & $11.177 \pm 37.138$ & $1\,623$ & $12.166 \pm 11.461$ & $1\,300$ \\ 
	$4$ & $2.106 \pm 0.130$ & $16.7$ & $11.762 \pm 37.171$ & $1\,624$ & $12.424 \pm 11.120$ & $1\,300$ \\ 
	$8$ & $4.222 \pm 0.238$ & $16.7$ & $12.041 \pm 36.312$ & $1\,631$ & $13.167 \pm 11.613$ & $1\,308$ \\
	$16$ & $8.358 \pm 0.644$ & $16.7$ & $13.451 \pm 37.415$ & $1\,633$ & $14.111 \pm 11.278$ & $1\,308$ \\ 
	$32$ & $16.787 \pm 1.03601$ & $16.7$ & $16.048 \pm 36.460$ & $1\,632$ & $16.661 \pm 11.122$ & $1\,308$ \\ 
	$64$ & $31.696 \pm 0.737$ & $16.8$ & $21.794 \pm 37.302$ & $1\,640$ & $22.189 \pm 11.531$ & $1\,316$ \\
	\hline
\end{tabular}
\end{table*}

\section{Designs Behind BurTorch's Low Latency}
\label{sec:design-for-low-latency}

\libname{BurTorch} excels in CPU gradient computation, especially for small batch sizes, making it ideal for memory-constrained scenarios. Next, we outline key design choices that contribute to its low latency. For further details, see Appendix \ref{app:design-philosophy}.

\paragraph{\color{black}Compile-time optimizations.}
\libname{BurTorch} maximizes compile-time optimizations usage, reducing runtime overhead by treating constants as compile-time values, thus optimizing execution. Unlike high-level frameworks, which limit such optimizations, \libname{BurTorch} leverages the full potential of modern CPUs, bypassing performance bottlenecks related to loops and memory access in all computation stacks. \libname{BurTorch} advocates for compile-time optimization across a fully integrated deep-learning training solution.

\paragraph{Eliminating unnecessary abstractions.}
Operating with no dependencies except for the OS, \libname{BurTorch} simplifies runtime execution. Through dead code elimination and whole-program optimization, compilers remove or simplify abstractions that exist only at design time. A small codebase offers several advantages. First, it results in a physically smaller compiled binary. Second, a smaller codebase allows for meticulous optimization of every detail coherently and holistically without neglecting other aspects of training or gradient computation—something that becomes practically \textit{impossible} with a large codebase.

\paragraph{\color{black}Instruction-level parallelism.} \libname{BurTorch} employs instruction-level parallelism by unrolling key operations like inner products, ensuring efficient use of CPU pipelines. By simplifying the compiler’s task, it helps maximizing performance, avoiding inefficiencies common in complex code structures. For further details, see Appendix \ref{app:compiler-in-general}.

\paragraph{\color{black}Efficient memory management.}

\libname{BurTorch} optimizes memory access by reducing latency, particularly when storing operands in DRAM. It stores partial derivatives and activations contiguously, minimizing fragmented memory access. \libname{BurTorch} introduces a "rewind" mechanism that discards unnecessary parts of the computation graph.

The ability to easily customize \libname{BurTorch} is also important. For example, in the Transformer block (Section \ref{sec:exp4-burtorch-fot-gpt3}), the concatenation of multiple self-attention heads typically involves memory copying in typical frameworks. \libname{BurTorch} avoids this by passing a sequence of memory views for linear layers, conceptually representing a split tensor without physically concatenating it. This approach eliminates unnecessary memory copies, which are approximately $\times 330$ more expensive compared to a single arithmetic operation if memory access happens to DRAM \citep{gregg2014systems}. This functionality is achievable with \libname{BurTorch}'s small codebase.

\paragraph{\color{black}Optimized backpropagation.}
\libname{BurTorch}'s backpropagation is optimized for latency. For details see Appendix~\ref{app:cap-burtorch-high-with-low}.

\section{Impact on Optimization Theory} 
\label{sec:inluence-on-theory}

\paragraph{\color{black}Practical implementations of theoretical algorithms.} \libname{BurTorch} optimizes the Backpropagation Algorithm through system-level improvements, efficient memory access, and a design that significantly reduces runtime overhead in both memory and computation.  It introduces a practically effective gradient oracle implementation  $\nabla f(x) = \nicefrac{1}{b} \sum_{i=1}^{b} \nabla f_i(x)$ which outperforms all state-of-the-art practical solutions for small batch sizes $b$ or low-dimensional settings $d$.

First, $b = 1$ is theoretically optimal for non-convex algorithms like \algname{PAGE} \citep{li2021page, tyurin2022sharper}. Before \libname{BurTorch}, achieving such low-latency gradient computation was impractical, limiting \algname{PAGE} applicability. \libname{BurTorch} removes this barrier, making \algname{PAGE} more practical. 

Second, in finite-sum convex optimization, where \algname{Proximal Stochastic Gradient Descent} with \algname{SGD-NICE} subsampling requires an optimal batch size of $\tau \approx 1$ to maintain low gradient variance \citep{Gower2019}, \libname{BurTorch} enables efficient computation of the subsampled gradient $\hat{\nabla f(x)}$.

\libname{BurTorch} demonstrates that system-level optimizations, such as refining memory access patterns, adopting non-recursive computation, and reusing memory buffers, have a significant impact on performance. These optimizations dramatically reduce hidden constants in $\mathcal{O}$ notation, offering a new perspective on optimization algorithms in the context of modern hardware constraints. Notably, DRAM access latency is approximately $330\times$ slower than register access \citep{gregg2014systems}, underscoring the importance of memory access considerations in the theoretical optimization algorithms.

\paragraph{Modifying the notion of the gradient oracle.}

Some optimization algorithms enable the interleaving of gradient computation, compression, and information transfer \citet{burlachenko2023federated}. However, implementing these techniques effectively in reality requires significant effort to seamlessly integrate communication and computation within Automatic Differentiation (AD). Research on asynchronous \algname{SGD} methods \citep{maranjyan2024mindflayer, maranjyan2025ringmasterasgdasynchronoussgd} further necessitates the incorporation of early termination—the ability to halt the computation of $\nabla f(x)$ upon request. However, even minor modifications to backpropagation often lack practical implementations in academic settings due to the inherent complexity of conventional frameworks. Maximizing throughput frequently results in overly intricate and non-maintainable implementations. \libname{BurTorch} simplifies these processes due to its compact design.

\paragraph{Randomized backpropagation.}

Early work by \citet{oktay2020randomized} investigated a randomization technique within backpropagation, but its full potential remains largely unexplored. The scalar-level granularity of \libname{BurTorch} allows these randomization techniques to be directly \textit{implemented} by modifying its codebase, rather than introducing additional layers to \textit{simulate} randomization.

\paragraph{Practical refinements for theoretical algorithms.}

A significant research direction in Federated Learning focuses on compression techniques that apply compression operations to gradient-like quantities. These methods use specialized operators, such as $\mathcal{C}(\hat{\nabla f(x)} - g)$ in \algname{EF21} by \citet{richtarik2024error} and $\mathcal{C}(\hat{\nabla f(x)} - \hat{\nabla f(y)})$ in \algname{MARINA} by \citet{gorbunov2021marina}, where $\mathcal{C}: \RD \to \RD$ is deterministic or randomized mapping. By design, these compression operators enable efficient transmission of compressed information over communication networks. For a survey on compression operators, see \citet{beznosikov2020biased}. For optimization algorithms that require efficient computation of the gradients \(\nabla f(x)\) and \(\nabla f(y)\) at two different iterates \(x, y \in \RD\), \libname{BurTorch} provides this functionality effectively out of the box. By leveraging low-level optimizations with {SIMD registers}, \libname{BurTorch} ensures high-performance execution natively.

\paragraph{Refining gradient compression to partial derivative granularity.}

Some compression methods \citep{beznosikov2020biased} operate independently of evaluated $\nabla f(x)$, meaning they do not require prior knowledge of $\nabla f(x)$ before being applied. \compname{RandK} sparsification, which selects a subset of $[d]$ coordinates independently of $\nabla f(x)$, is one such example. 

The time complexity of computing $\frac{\partial f(x)}{\partial x_i}$ depends on the location of $x_i$ within the computation graph. If $x_i$ is in the early layers of a DL model, both forward and backward passes are required. In contrast, if $x_i$ is in the later layers, only a forward pass and a single backward step are needed, avoiding traversal of the entire model during backward. When computing $\frac{\partial f}{\partial \textbf{z}}$ for $\textbf{z} \in \{x_1, \dots, x_k\}$, the computational cost depends on their positions in the computation graph. This idea can be leveraged to design more efficient compression techniques. The \compname{RandSeqK} compressor \citep{burlachenko2024unlocking} was developed to group spatially close coordinates, optimizing memory access through coalesced memory operations. A similar strategy could be employed by exploiting the internal flexibility of the \compname{RandK} compressor to create structure-aware compression methods for $f(x)$. The fast oracle in \libname{BurTorch} enables computation of $\nabla f(x)$ restricted to specific coordinate subset $S$, facilitating the development of either exact (for small $d$) or approximate cost models (for big $d$) to evaluate $[\nabla f(x)]_{i,i \in S}$.

\paragraph{Coupling gradient computation with sparsification.}

While gradient sparsification can reduce communication and storage overheads by decreasing the number of non-zero gradients, modern Deep Learning frameworks such as \libname{JAX}, \libname{TensorFlow}, and \libname{PyTorch} do not natively offer direct support for sparse gradients or gradients with specific structures. In contrast, \libname{BurTorch}, with its transparent architecture, can be conceptually adapted to support gradient sparsification, as long as the sparsification rule is based solely on individual $\frac{\partial f_i(x)}{\partial x_j}$, and not on the entire gradient $\nabla f_i(x)$ or $\nabla f(x)$.

\section{Conclusions}
\label{sec:conclusion}

We introduced \libname{BurTorch}, a CPU-based backpropagation implementation. Real-world experiments conducted across Windows (Section~\ref{sec:experiments}), Linux (Appendix~\ref{app:extra-experiments-linux}), and macOS (Appendix~\ref{app:extra-experiments-macos}) on various devices demonstrate significant improvements in computation latency, memory usage, and energy efficiency (Appendix~\ref{app:energy-eff}) during the computation of $\nabla f(x)$. \libname{BurTorch} has been compared against \libname{JAX}, \libname{TensorFlow}, \libname{PyTorch}, \libname{Apple MLX}, \libname{Autograd}, \libname{TFLite}, and \libname{Micrograd}. The compact codebase of \libname{BurTorch} lays the foundation for adapting optimization techniques to backpropagation and tailoring backpropagation to specific optimization algorithms in scenarios where $\nabla f(x)$ is computed through backpropagation.

\clearpage
\bibliography{burtorch-icml25-after}
\bibliographystyle{icml2025}


\clearpage
\onecolumn
\tableofcontents

\clearpage
\appendix

\section{Missing Details for Experimental Setup}
\label{app:exp-setup}

\subsection{Hardware and software environment for Windows OS experiments}
\label{app:exp-setup-win}

\begin{enumerate}
\item Operating System: Microsoft Windows 11 Home.
\item CPU: Intel Core Ultra 7 155H, x86-64, Little-Endian.
\item CPU Clock Frequency: $4.48$ GHz during experiments (via \texttt{powercfg.cpl})
\item CPU Cores: 22 Logical, 16 Physical. During the experiments, only one core was used.
\item Physical Memory: 64 GBytes, DDR5, $2792$ MHz.
\item Hard Drive: KXG80 NVMe SSD, Sector size: 512 bytes, Filesystem: NTFS
\item Python Interpreter: Python: 3.9.0,
\item Python Libraries: TensorFlow: 2.8.0, Autograd 1.7.0, PyTorch: 2.5.1, JAX: 0.4.30, Micrograd: c911406, NumPy: 1.23.0, TensorFlow Lite (the same TensorFlow version): 2.8.0. C++ Libraries: LibTorch (C++ \libname{PyTorch} API): 2.5.1
\item {BurTorch} and LibTorch examples built with: Microsoft Visual Studio 2020 v17.11.4 (MSVC 14.41.34120) and \texttt{/O2,/Oi,/GL}, CMake version: 3.30.4
\end{enumerate}

\subsection{Hardware and software environment for Linux OS experiments}
\label{app:exp-setup-linux}

\begin{enumerate}
\item Operating System: Ubuntu 20.04.6 LTS.
\item CPU: Intel(R) Xeon(R) Gold 6146 CPU, x86-64, Little-Endian.
\item CPU Clock Frequency: $3.2$ GHz during experiments.
\item CPU Cores: 24 Logical, 24 Physical. During the experiments, only one core was used.
\item Physical Memory: 251 GBytes, DDR4, $2666$ MHz.
\item Hard Drive: Seagate ST4000NM0035, HDD, Sector size: 512 bytes, Filesystem: EXT4.
\item Python Interpreter: Python: 3.9.21.
\item Python Libraries: TensorFlow: 2.18.0, Autograd 1.7.0, PyTorch: 2.5.1, JAX: 0.4.30, Apple MLX: 0.22.0, Micrograd: c911406, NumPy: 2.0.2. LibTorch (C++ {PyTorch} API): 2.5.1.
\item {BurTorch} and LibTorch examples was build with g++-11 and \texttt{-O3 -flto} compilers flags, CMake version: 3.27.
\end{enumerate}

\subsection{Hardware and software environment for macOS experiments}
\label{app:exp-setup-macos}

\begin{enumerate}
\item Operating System: Sonoma 14.5.
\item CPU: Intel Quad-Core Intel Core i7, x86-64, Little Endian.
\item CPU Clock Frequency: $2.3$ GHz during experiments.
\item CPU Cores: 8 Logical, 4 Physical. During the experiments, only one core was used.
\item Physical Memory: 32 GBytes, LPDDR4X, $3733$ MHz.
\item Hard Drive: Macintosh HD APPLE SSD AP2048N, SSD, Sector size: 4096 bytes, Filesystem: APFS.
\item Python Interpreter: Python: 3.9.21.
\item Python Libraries: TensorFlow: 2.16.2, Autograd 1.7.0, PyTorch: 2.2.2, JAX: 0.4.30, Apple MLX: 0.7.0, Micrograd: c911406. LibTorch (C++ {PyTorch} API): 2.2.2.
\item {BurTorch} and LibTorch examples was build with Apple CLang 15.0.0 and \texttt{-O3 -flto} compilers flags, CMake version: 3.27.
\end{enumerate}

\clearpage
\section{Discussion on Compiler Considerations}
\label{app:compiler-in-general}

\subsection{On why a compile-based approach is fundamentally different from utilizing scripting languages}
\label{app:why-compilers}

As noted in \citet{wexelblat2014history}, compiler research dates back to the 1950s, with John Backus’s Speedcoding project addressing rising software costs. While compiled languages are often considered low-level in modern ML research, scripting languages, due to their higher levels of abstraction, can obscure essential control over interactions with hardware and the OS. Understanding why a compiler-based approach is superior requires examining the inner workings of a compiler system. A compiler operates in several distinct stages:

\begin{enumerate} 
\item \textbf{Lexical analysis.} The implementation's source code is divided into tokens which are the smallest indivisible units within a language, such as keywords, operators, separators, identifiers, and literal constants. These symbols form the foundation for all subsequent analysis.

\item \textbf{Syntax analysis.}  The compiler constructs an Abstract Syntax Tree (AST), validating that the program adheres to the language’s syntactic rules, defined typically using Backus–Naur forms for context-free grammars (CFG).

\item \textbf{Semantic analysis.} This phase ensures that the program is semantically correct, checking type compatibility, variable scope, and other areas that syntax analysis alone cannot address.

\item \textbf{Code optimization.} Code optimization improves runtime performance and reduces memory usage in various forms.

\item \textbf{Code generation.} Finally, the compiler generates final machine instructions, directly producing binary code for the target architecture or intermediate assembly code, which is then converted into the machine code.

\end{enumerate}

In contrast, interpreters execute scripts directly at runtime, without preprocessing. Therefore interpreters typically introduce significant overhead by bridging abstractions during execution. This results in inefficiency in runtime, especially in large-scale or real-time systems like ML, where performance is critical. 

To address these challenges, there are efforts from scripting-based approaches:

\begin{itemize} 
\item Establishing best practices for specific scripting languages.
\item Developing specialized libraries or frameworks for performance improvement.
\item Augmenting the scripting ecosystem with just-in-time (JIT) compilers.
\item Improving runtime efficiency through external tools or hybrid approaches.
\end{itemize}

However, in our vision, adopting a compiler-based approach in ML research presents a significantly more efficient and scalable alternative. By leveraging the power of compilers, researchers can achieve superior performance, direct hardware control, and enhanced scalability—eliminating the inefficiencies and compromises inherent in interpreted languages.

Historically, this approach has not been widely adopted, primarily because large codebases make compilation slow, creating a social barrier to its use. \libname{BurTorch}'s compact code design overcomes this limitation, making compiler-based ML development both practical and efficient.

\subsection{On performance optimizations achievable through compiler utilization}
\label{app:what-compilers-can}

At the \textbf{Optimization} stage, the Abstract Syntax Tree (AST) for the program has been constructed and augmented with information derived from the semantic analysis phase. Compiler optimizations are rooted in a broad array of innovations from various scientific and engineering disciplines, which, when applied effectively, yield substantial performance improvements. To try to optimize computation time, the compilers generally execute a sequence of transformation passes, each of which analyzes and refines the code to optimize its performance. Each transformation pass may iterate multiple times, with the steps typically following a predetermined order that has been demonstrated empirically effective in most cases. The following summarizes common optimization techniques:

\begin{itemize}
\item \textbf{Arithmetic simplification.} Optimization by converting complex arithmetic operations into less expensive ones using bitwise operations and shifts with binary numbers.

\item \textbf{Register allocation.} The substitution of stack-based storage with processor registers, improves data access speeds.

\item \textbf{Memory layout optimization.} Rearranging the layout of data structures, to improve memory access patterns.

\item \textbf{Data structures transformation.} Modifying the layout of data structures to maximize the use of registers.

\item \textbf{Dead code elimination.} Removal of unreachable or redundant code from the program’s control flow.

\item \textbf{Function inlining.} The compiler employs heuristics to decide which functions should be inlined, i.e., the invoking of a function is replaced with inserting the body of the function directly into the implementation.

\item \textbf{Hoisting.} Moving loop-invariant code outside the loop to reduce redundant computation.

\item \textbf{Loop unrolling.} Unrolling loops to reduce control overhead and potentially exploit Instruction Level Parallelism.

\item \textbf{Loop fusion or loop jamming.} Merging multiple loops that iterate over the same range of indices.

\item \textbf{Register allocation policies.} Specialized heuristics to determine optimal register usage during runtime.

\item \textbf{Constant propagation.} The propagation of constant values throughout the program’s code, enabling compile-time evaluation of expressions and eliminating computing or fetching them from memory in runtime.

\item \textbf{Global program optimizations.} The entire implementation is treated as a unified system, allowing for deeper optimization by leveraging additional knowledge such as bypassing standard calling conventions between subroutines.

\end{itemize}

\subsection{On performance optimizations beyond traditional compiler capabilities}
\label{app:what-compilers-can-not}

Compilers automate the translation of code between languages, optimize execution through iterative transformations, and ensure correctness at the programming language level. While they can handle many optimizations, several aspects fall outside their scope: (i) an algorithm's design; (ii) specific implementation choices; (iii) algebraic invariants.

These limitations have persisted for decades, though future advancements may enable compilers to incorporate more invariance-aware optimizations. Specifically, the following limitations remain:

\begin{itemize} 

\item Compilers cannot solve undecidable computational problems, such as Halting Problem~\citep{leiserson2020there}.

\item Constructing cache-oblivious algorithms (those that automatically adapt to multilevel cache hierarchies and their sizes) is a complex task that requires manual design~\citep{demaine2002cache}.

\item Automatically selecting the most efficient form of disk access for a given situation. For example, when reading datasets from disk, the optimal access method is often counterintuitive—subtle OS interfaces like \texttt{mmap} may outperform traditional read/write APIs~\citep{burlachenko2024unlocking}. 

\item Providing formal guarantees that generated code is optimal. 

\item Optimizing across multiple programming environments (for example, efficiently integrating C++ and Python). 

\item Redesigning data structures to improve spatial or temporal locality. 

\item Performing early-exit tests, such as determining whether a point lies outside a complex polyhedron using precomputed bounding box information.\footnote{See lecture slides from MIT 6.172: Performance Engineering of Software Systems by C. Leiserson and J. Shun.}

\item Implementing fundamental algorithmic improvements and eliminating semantically redundant computations beyond standard optimizations like dead code elimination and common subexpression elimination. 

\item Coarsening recursion. The concept of coarsening recursion involves constructing a base-case algorithm that effectively handles small tasks, despite poor asymptotic behavior for large input sizes, and using this algorithm within the structure of a larger computational problem.

\item Some aspects of inlining. Inline subroutine calls are useful not only for eliminating call overhead but also for opening opportunities for further improvements. However, compilers, even when forced to inline subroutines physically, sometimes face challenges in inlining functions. A notable example is recursive functions\footnote{For instance, constraints on inlining in the Microsoft MSVC Compiler can be found here: \href{https://learn.microsoft.com/en-us/cpp/cpp/inline-functions-cpp?view=msvc-170}{https://learn.microsoft.com/en-us/cpp/cpp/inline-functions-cpp?view=msvc-170}}.
\end{itemize}

Expanding compiler optimizations beyond their traditional scope is an active area of research in both the compiler community and broader fields dealing with algorithmic logic. Examples include research on extending algebraic expression replacement beyond simple transformations~\citep{jia2019taso}, automatic function generation for inference~\citep{wu2024mirage, fawzi2022discovering}, and the discovery of fundamentally new algorithms~\citep{fawzi2022discovering}.

Despite these advancements, an alternative approach with significant potential exists. By focusing on a specialized class of models (for example, Transformer-based architectures), a specific class of optimization algorithms (for example, first-order methods), and a dedicated execution strategy for gradient computations (for example, backpropagation), many applications can be expressed within these constructs. In such cases, the full generality of current compiler research may not be necessary, as many limitations can be addressed manually.

\subsection{On compilation techniques for enhancing {PyTorch} backpropagation speed on CPU for training}

\label{app:torch-compile-techniques-backprop-speed-cpu}

To the best of our knowledge, both in academia and industry, \libname{PyTorch} \citep{paszke2019pytorch} is predominantly used through its Python API. When improving backpropagation speed is critical, \libname{PyTorch} offers several optimization methods.

\paragraph{{PyTorch} graph execution mode via TorchScript.}

One of the key differences between \libname{PyTorch} and \libname{TensorFlow 1.0} lies in their execution strategies. While \libname{TensorFlow 1.0} uses a static graph-based execution model, \libname{PyTorch} adopts dynamic, eager execution, which allows for more intuitive and flexible computation. \libname{TensorFlow 2.0} later introduced eager execution (see Section~\ref{sec:prev-systems}), while maintaining backward compatibility with \libname{TensorFlow 1.0} graph mode, as static computation graphs offer advantages in execution speed and optimization opportunities. To leverage this model, \libname{PyTorch} introduced TorchScript, a tool that converts dynamic, eager code into a static, graph-based representation (with some limitations). In our experiments, we utilize TorchScript through \libname{PyTorch's} built-in tracing capabilities to achieve performance improvements.

\paragraph{{PyTorch} LibTorch C++ API.}

In addition to TorchScript, another approach for improving the performance of PyTorch-based algorithms is through \libname{PyTorch’s} native C++ interface, called LibTorch. LibTorch allows direct integration of \libname{PyTorch} into C++, offering an alternative to the Python API. At its core, LibTorch includes the ATen library, which serves as the foundation for tensor operations and Automatic Differentiation. The C++ API closely mirrors its Python counterpart, ensuring a consistent and familiar experience for developers transitioning between the two environments. Our experiments demonstrate that, unfortunately, switching from Python to C++ does not automatically unlock all the advantages of a compiler-based approach. This is partly due to the large codebase of PyTorch.

\paragraph{CPU vendor-specific extensions for PyTorch.}

The release builds of \libname{PyTorch} include functionality for specific computations (backends). However, in some cases, CPU vendors may extend \libname{PyTorch} in different ways to optimize performance on their specific hardware platforms. These extensions can further accelerate backpropagation speed by utilizing CPU-specific optimizations tailored to the vendor's architecture. To use these extensions, users typically need a specific version of \libname{PyTorch} and must define certain environment variables. By default, \libname{PyTorch} attempts to use and is compiled with the best-practice compute libraries.

\subsection{On execution modes} 
\label{app:exex-details-of-comp}

\paragraph{Eager mode.} In our experiment,  \libname{Apple MLX}, \libname{Autograd}, \libname{Micrograd}, \libname{BurTorch}, and certain configurations of \libname{PyTorch} and \libname{TensorFlow} operate with a dynamic computation graph, where forward operations are executed immediately as they are encountered in the description. This model of execution is referred to as \textit{Eager Mode}. In this mode, there is no graph compilation process, and no operation tracing for analysis or conversion takes place.

Interestingly, \libname{BurTorch} outperforms graph-based frameworks despite operating in eager mode. Our experiments show that eager execution can be faster than graph mode in certain situations, demonstrating the advantages of \libname{BurTorch} in handling such cases efficiently.

\paragraph{Graph mode.} In our experiment, we utilized \libname{TensorFlow}, \libname{TF Lite}, and \libname{PyTorch} in a Just-In-Time (JIT) manner to compile and generate a static computation graph that represents $f(x)$. This execution model is referred to as \textit{Graph Mode} and is the modern approach for constructing computation graphs, in contrast to the manual construction required in the past (see \libname{TensorFlow 1.0} in Section~\ref{sec:prev-systems}). 

In this model, the $f(x)$ computation is represented in a framework-specific manner, with the framework's runtime tracing the execution of operations. Upon the first call of $f(x)$, the internal compiler converts the function operations into a graph form. After this conversion, subsequent computations become more efficient.


\clearpage
\section{Discussion on Backpropagation}
\label{app:backprop}

\subsection{On backpropagation memory taxonomy}
\label{sec:backprop-memory}

In supervised ML tasks, the score function is typically organized according to the principle of Empirical Risk Minimization. Modern computational frameworks, such as \libname{TensorFlow} \citep{abadi2016tensorflow} and \libname{PyTorch} \citep{paszke2019pytorch}, enable the automated computation of gradients with respect to trainable weights, also known as optimization variables. These gradients are evaluated on complex and arbitrarily structured computational graphs using algorithms designed for efficient derivative computation. This process is primarily carried out through backpropagation, a form of Automatic Differentiation in Reverse Accumulation Mode~\citep{griewank2008evaluating}.

Let us assume we use {backpropagation} for a model containing $d$ trainable variables and $d' \ge d$ total variables, which include both trainable and non-trainable parameters, along with a batch of $b$ input-output pairs. The distinction between $d$ and $d'$ arises when only a subset of the model's parameters is trained, as in fine-tuning and transfer learning.

When switching from the \texttt{Forward Pass} to the \texttt{Backward Pass}, the {backpropagation} conceptually requires memory storage for the following quantities:

\begin{enumerate} 
\item \textit{Trainable parameters or variables (weights)}. Memory footprint: $d$ scalars. \newline 
Encodes the trainable part of the Deep Learning model.

\item \textit{Non-trainable parameters or non-trainable variables (frozen weights)}. Memory footprint: $d' - d$ scalars. \newline 
Represents the portion of the model not involved in training, comprising $d' - d$ scalars.

\item \textit{Activations or activation maps (response maps)}. Memory footprint: $d' \times b$ scalars. \newline
Output for all neurons (computational nodes, graph operators).

\item \textit{The $(\mathrm{Input}_i, \mathrm{Output}_i)$ pair for each trainable sample $i \in [b]$}. Memory footprint: $b \times$ \textit{"Memory for Single Input"} + $b \times$ \textit{"Memory for Single Output"}. \newline
The input part is used only in the first step of the \texttt{Forward Pass} and in the last step of the \texttt{Backward Pass}. The output part is used in the last step of the \texttt{Forward Pass} and the first step of the \texttt{Backward Pass}. If computation is not done on the CPU, transferring $(\mathrm{Input}_i, \mathrm{Output}_i)$ pairs to the computational device may become a bottleneck, especially when the descriptions of these pairs are large or when the communication bus is highly contended. For instance, transferring data to a GPU connected via a PCI-Express bus may require additional considerations. In such cases, systems like \libname{nvCOMP}~\citep{nvcomp} can optimize data transfer workflows.

\item \textit{Error signal from connected layers}. Memory footprint: $2 \times$ the maximum number of single activations in the DL model layer. \newline        
The {backpropagation} algorithm does not explicitly store Jacobians; instead, it implicitly evaluates them through the propagation of error signals, typically denoted by $\delta$. Error propagation begins as soon as the \texttt{Backward Pass} starts. The peak memory for storing $\delta$ is proportional to the maximum number of compute units in two consecutive layers.

\item \textit{Optimizer state}. Memory footprint: depends on the optimizer. \newline
The memory footprint depends on the optimizer. \algname{Gradient Descent} has a footprint of $0$. In popular optimizers like Adaptive Moment Estimation ({ADAM}), the footprint is $2d$ scalars, which track the element-wise gradient direction and the element-wise squares of the $\nabla f(x)$. Work addressing this challenge includes {MicroAdam}~\citep{modoranu2024microadam}.

\end{enumerate}

\subsection{On activation maps memory footprint discussion and solutions suggested via BurTorch} \label{sec:resolve-mem-activations-via-lat-design}

Let us now focus on the memory footprint for activations. If all operators in the compute graph are considered as single activation functions $\mathbb{R} \to \mathbb{R}$, the memory footprint for activations in throughput-oriented DL systems is $d' \times b$ scalars.

The multiplicative factor of $d'$ arises because, even if a trainable variable is located in the first layer, $d'$ can still be significantly larger than $d$. The batch size multiplier emerges due to the lack of activation sharing across samples in the subsampled batch of size $b$. Addressing this challenge remains an open research question. One promising approach is Activation Compression, which has shown potential in reducing training memory usage \citep{liu2022gact}. Another technique is gradient checkpointing~\citep{griewank2008evaluating,chen2016training}. The multiplicative factor $d' \cdot b$ often leads to memory bottlenecks in throughput-oriented designs during {backpropagation}. This issue arises both in training Convolutional Neural Networks\citep{mishra2017wrpn} and in full training and fine-tuning of Large Language Models~\citep{2024fwdllm}.

\libname{BurTorch} offers an alternative solution by addressing the root causes of this problem. The issue primarily arises in throughput-oriented designs, which focus on maximizing the parallel processing of tasks. \libname{BurTorch}, being a latency-oriented design, focuses on optimizing the processing of computations in a serialized manner. This approach can be particularly effective in situations where computations, such as the calculation of sample-gradient oracles, are performed sequentially, thus alleviating the memory bottleneck associated with activation storage in throughput-oriented systems.

\clearpage
\section{Discussion on Small Compute Graphs}
\label{app:on-small-compute-graphs}

There are several compelling reasons to emphasize the importance of small compute graphs, even though modern DL is more focused on big dimensional problems.

\paragraph{Mathematical standpoint.}

From a mathematical perspective, the classical Cantor diagonal theorem tells us that $\mathbb{R}^d \sim \mathbb{R} \sim [0,1]$, meaning that the line segment $[0,1]$ contains the same number of points as $\mathbb{R}^d$, as there exists a bijection between them. While this is theoretically interesting, it has limited practical implications. In real-world implementations with finite precision, approximations of $\mathbb{R}$ can accommodate no more points than $\mathbb{R}^d$, highlighting the constraints imposed by finite precision in practical systems.

Next, the Kolmogorov-Arnold Theorem (KAT) asserts that any continuous function $f: \mathbb{R}^d \to \mathbb{R}$ can be represented as:
\[
f(x) = \sum_{j=1}^{2d+1} g_{f,j} \left( \sum_{i=1}^d \lambda_i \gamma_i(x_i) \right),
\]
where $\gamma_i: \mathbb{R} \to \mathbb{R}$ and $\lambda_i \in \mathbb{R}$ are independent of $f$, and $g_{f,j}: \mathbb{R} \to \mathbb{R}$ is entirely determined by $f$. The KAT theorem redistributes the complexity of $f(x)$ into scalar functions $g_{f,j}$, demonstrating that the number of variables $d$ does not always serve as an adequate measure of complexity \citep{lorentz1966approximation}.

\paragraph{Discrete optimization tractability for small ML models.} In Machine Learning, particularly Deep Learning, incorporating prior knowledge often involves selecting a class of parameterized functions, applying regularization schemes for optimization, and using robust training techniques. While small compute graphs may restrict modeling flexibility, they offer advantages in tasks with a combinatorial nature. For instance, small compute graphs can be beneficial for identifying globally optimal pruned models or when certain parts of the trainable variables are constrained to small discrete sets. Discrete Optimization, when applied without heuristics, faces fundamental limitations in high-dimensional problems. However, when the number of discrete variables is small, the problem becomes more manageable.

\paragraph{Applications of on-device training on CPU.} Small compute graphs are particularly essential for resource-constrained systems, such as Internet of Things (IoT) devices, which often face limitations in computational power, memory, and energy resources. By reducing memory and computational demands, small compute graphs are well-suited for these environments. Furthermore, many IoT applications such as real-time monitoring, autonomous systems, and smart sensors—require low-latency processing. 

Smaller compute graphs enable faster inference, facilitating real-time decision-making while minimizing energy consumption. In the context of on-device training, Deep Learning models can enhance or even replace traditional processing pipelines. The engineered systems developed in recent decades present substantial opportunities for optimization across the computational stack. However, to maintain practical performance, on-device training must operate on millisecond timescales for real-time applications (for example, graphics, gaming, and video streaming) and nanosecond timescales (network protocols). 

From a hardware perspective, it is extremely challenging for CPUs to compete with GPUs in terms of processing throughput. However, it is important to note that most practical \textit{applications} are inherently implemented on CPUs. Therefore, for CPU vendors to remain competitive in the domain of practical Deep Learning systems, it may be beneficial to focus on latency-sensitive use cases in on-device training. \libname{BurTorch}, with its latency-optimized design, plays a critical role in enabling traditional applications, typically implemented as compute programs, to leverage on-device training concepts in scenarios such as Federated Learning.

\clearpage
\section{Design Philosophy Behind BurTorch}
\label{app:design-philosophy}


The design of \libname{BurTorch} prioritizes simplicity in development, debugging, and optimization while maintaining high performance. Its architecture and implementation is guided by the following principles:

\paragraph{\color{black}{1: Leveraging a compile-based Language from the ground up.}} {\libname{BurTorch}} is built using modern C++20, a widely supported language standard that aligns with the capabilities of current compilation tools. This foundation enables \libname{BurTorch} to significantly reduce the latency of gradient computation, especially for small batch sizes, by utilizing templates to generate only the necessary algorithms and code at compile time. This approach ensures maximum efficiency while minimizing runtime overhead, eliminating the need for dynamic dispatch, which is prevalent in languages like Java, C$\#$, and Python. 

While C++ is not traditionally considered the primary language for research prototyping due to its emphasis on standardization and perfection, this focus can create high barriers to entry, slowing rapid experimentation and iteration. In contrast, prototyping demands flexibility and speed, attributes that Python excels in. The philosophy behind \libname{BurTorch} is that, despite C++'s complexity, the codebase remains small and clean, and the user experience, when using compile-based techniques, is indistinguishable from that of script-based languages, as compilation time is negligible in this case.

\paragraph{2: Leverage efficient looping and function invocation.} The design philosophy behind \libname{BurTorch} is rooted in the principle that when input organization necessitates loops, we prioritize implementing them directly in the code. While Python-based frameworks offer various indexing mechanisms that provide flexibility, this flexibility comes with trade-offs. Our observations suggest that manually implementing the necessary functionality often yields better results. In manually created loops, which remain maintainable, redundant computations can be effectively removed through various optimization strategies. 

Furthermore, manually created loops—while offering comparable computational capabilities—serve as a valuable educational tool, providing users with greater control over the computational process. In contrast, users relying on interpreted environments often face limitations, as several nested loops cannot be efficiently implemented at the script level of Python. It is not only loops that cannot be implemented efficiently but also function dispatching and attribute access mechanisms. This is because, in Python, method and data attribute access are dynamically bound and resolved at runtime.\footnote{As stated in all official Python tutorials from \href{https://www.python.org}{www.python.org} up to and including Python 3.11.}

\paragraph{3: One language and compact code for debugging and optimization.}

Managing and debugging systems implemented in multiple languages is inherently challenging. Python-based systems, for example, face three key issues: (i) debugging across different languages is challenging, (ii) building extension modules with debug info as the number of used libraries grows is challenging, and (iii) understanding code intent in large codebases, especially when performance optimizations sacrifice clarity is challenging for everybody (including authors). These challenges hinder holistic optimizations. \libname{BurTorch} overcomes these issues by maintaining a small codebase and using a single language (C++20), which simplifies debugging and profiling from high-level design to scalar operations. This approach ensures streamlined debugging and is most effective when Deep Learning models are built from simple components with clear training algorithms.

\paragraph{4: Provide high-level and low-level abstractions with minimal overhead.}

A key design principle of \libname{BurTorch} is its ability to offer both high-level abstractions for convenience and low-level control for performance, with minimal overhead connecting the two. High-level abstractions may sacrifice performance, primarily due to two factors.

First, they often lack customization for critical operations. For instance, many frameworks introduce unnecessary memory copying during tensor concatenation \footnote{Concatenate tensors without memory copying \href{https://discuss.pytorch.org/t/concatenate-tensors-without-memory-copying/34609/8}{https://discuss.pytorch.org/t/concatenate-tensors-without-memory-copying/34609/8}}, which can degrade efficiency. High-level languages make it difficult to identify such low-level inefficiencies. But optimizing such things is crucial. For memory operations accessing bytes from DRAM is $\times 330$ more latency-sensitive than scalar arithmetic operations \citep{gregg2014systems}. \libname{BurTorch} delegates performance optimization to the specific implementation of the compute graph, ensuring that core computations are performed with minimal overhead.

The second factor involves the role of \textit{interfaces}. Interfaces bridge different systems, components, or fields. However, in performance-critical situations, these interfaces can introduce significant overhead. Transferring between languages and making function calls can add latency, and without measuring system performance holistically, even simple implementations can outperform complex state-of-the-art systems \citep{mcsherry2015scalability}. \libname{BurTorch} minimizes this issue by leveraging compile-time optimizations such as code inlining, C++ template generation, global program optimization, and static builds from source code. This approach decouples the interfaces used for describing algorithms from those used at runtime.

\paragraph{\color{black}{5: Self-contained design.}} \libname{BurTorch} adopts a self-contained design idea that leverages standardized operating system interfaces and fully utilizes available computational devices without relying on external runtime systems or computation libraries. This approach minimizes overhead introduced by control structures and environments, which can significantly impact computation time. Operating entirely in userspace, \libname{BurTorch} is optimized for modern OS environments. Based on our experience, implementing functionality in kernel space offers limited benefits compared to the complexity and effort involved in handling compute-intensive and I/O-bound tasks. By remaining in userspace, \libname{BurTorch} streamlines development while ensuring high performance. This self-contained design also facilitates seamless integration into resource-constrained systems and removes the need to manage external library dependencies.

\paragraph{6: Addressing control overhead in complex frameworks.}

We categorize the control overhead in system implementations into two distinct types: unavoidable and avoidable. Unavoidable overhead arises from design choices in closed-source or overly complex open-source frameworks, where users have limited control over the internal architecture. These high-level abstractions obscure crucial details of the system, leading to inefficiencies in optimizing computational resources. In contrast, avoidable overhead results from design decisions in a specific implementation of some DL framework, which prioritize ease of use and modification over peak performance.

While Deep Learning frameworks provide useful abstractions, they often sacrifice mathematical rigor and computational efficiency. For instance, many frameworks and even the Python interpreter rely on pre-built dynamic libraries that cannot be selectively excluded during runtime. This forces the OS to load unnecessary operations, resulting in performance degradation. To address these challenges, \libname{BurTorch} adopts a minimalist, system-oriented approach: (i) prioritizing compile-time optimizations via native language compilers, (ii) exposing only essential compute and OS primitives for each task, (iii) offering simplified, user-friendly views of hardware and OS details, and (iv) maintaining a compact and efficient architecture.  This design philosophy reflects \libname{BurTorch}’s commitment to minimizing overhead at all software levels, from high-level to low-level interactions.

\paragraph{\color{black}{7: Relying on the static linkage of the final executable.}}

\libname{BurTorch} philosophy avoids reliance on dynamic libraries (except those required by the OS for invocation). This approach enhances portability and reduces cascading dependencies on middleware that are often inherited when using Python. By eliminating the need to prepare environments and download dependencies, this design allows for whole-program optimization at compile time. This process constructs native CPU code that can bypass traditional calling conventions, eliminating subtle misinterpretations of compute function interfaces.

Technically, whole-program optimization is nearly infeasible when each piece of logic is implemented as a standalone distributed library. In this context, the Python interpreter serves as an ecosystem, linking its runtime with Python extension modules. Systems built on these principles are inherently challenging to optimize through whole-program optimization, as the final binaries exhibit a more compact structure compared to their original source code design. The result is a streamlined implementation with optimized performance in \libname{BurTorch}.

\paragraph{8: Recognizing the importance of visualization in model development.}

Practical ML projects generally progress through the following phases:

\begin{enumerate}
\item Selection of the problem to address.
\item Collection of data, either through statistical tests or the problem's setup.
\item Design of a mathematical model, with or without domain expertise.
\item Training of the model, with or without guarantees of convergence.
\item Evaluation of the model's performance.
\item Deployment of the model for autonomous operation.
\item Ongoing maintenance, which may involve revisiting earlier phases.
\end{enumerate}

While this classical approach persists, new methodologies are emerging with the rise of large language models (LLMs). Python's widespread use in ML stems from its low entry barrier, essential in multidisciplinary fields. Additionally, Python’s rich ecosystem, including tools like Matplotlib, excels in data exploration and model design, aiding decision-making in phases (2) and (3). In contrast, general-purpose programming languages like C++ lack comparable visualization tools and have a higher entry barrier.

To bridge this gap, \libname{BurTorch}: (a) relies on core language constructs with a minimalist design; (b) provides an API similar to PyTorch; and (c) dynamically generates Python scripts to leverage tools like Matplotlib, potentially running separate processes during code debugging to facilitate visualization. 

This hybrid approach in \libname{BurTorch} supports the generation of Python scripts during debugging, enabling real-time plotting of scalar plots through a separate Python interpreter instance. Additionally, \libname{BurTorch} allows for visualization of computation graphs, which can be exported in DOT format for easy analysis. This replicates the familiar debugging and plotting workflows found in tools like Matlab or Python, enabling quick iteration during experimentation. By leveraging Python’s strengths in debugging and visualization, \libname{BurTorch} ensures flexibility and convenience during development while benefiting from C++'s computational efficiency and scalability for fast training execution. 

This hybrid model allows researchers to harness the best of both worlds: flexibility during debugging and performance during final runtime preparation. Importantly, \libname{BurTorch} does not attempt to integrate scientific visualization directly into the C++ runtime but instead acknowledges its essential role during model development and data exploration. By seamlessly integrating Python’s visualization tools with C++’s performance, \libname{BurTorch} facilitates the visualization of scalar metrics while maintaining high performance. A common question that may still arise is:

\begin{center}
\textit{Why doesn’t C++ have such rich libraries in the first place?}    
\end{center}

This is likely a result of the language's perfectionist principles, combined with the actual time required to create such libraries. The most notable of these principles, rooted in the language's origins \citep{stroustrup1994design}, are as follows:

\begin{center}
\textit{What you don’t use, you should not pay for. If something is built into the language itself, it cannot be done better by hand.} 
\end{center}


\paragraph{9: Addressing the limitations of gradient oracle interfaces.}

In supervised ML, Reinforcement Learning, Adaptive Control, and Optimization, the gradient computation process is typically automated through Automatic Differentiation (AD). Although AD simplifies gradient computation, it often lacks the flexibility required for tasks that require greater control over the gradient computation. Achieving practical, state-of-the-art performance requires more than automation; it requires precise control over each stage of computation. \libname{BurTorch} addresses this need through the following methodologies:

\begin{enumerate} 
\item A minimalistic codebase.
\item Leveraging a language tightly integrated with OS and CPU features.
\item A transparent, simple, and contiguous memory layout for storing partial derivatives, activations, and parameters.
\end{enumerate}

The memory regions storing activations and trainable variables in \libname{BurTorch} are contiguous, meaning they are not fragmented across different locations in the process's virtual memory. This is essential for tightly coupling training implementations with the underlying operating system and communication systems.

For example, the sequential nature of these buffers enables efficient read operations, optimizing model serialization to disk. Sequential reads and writes facilitate ideal CPU cache utilization (when not already occupied), and disk transfers are executed efficiently. It's important to note that both caches and disks do not operate at the byte level. All memory transitions between DRAM and CPU caches are managed by the DRAM Memory Controller, processing these transitions in fixed-size blocks known as \textit{cache lines}, typically 64 bytes long on most x86-64 and AArch64 architectures. Disk operations, on the other hand, occur at the sector level, usually 512 bytes.

Although networks and operating systems offer straightforward interfaces, it would be a mistake to simplify them to just basic mechanisms. A deeper understanding of these systems is essential for effectively coupling backpropagation computation with communication. A concrete example of an interface that benefits from contiguous memory is sending buffers to a Network Interface Controller (NIC) via the \texttt{MSG\_ZEROCOPY} mechanism\footnote{Available from \href{https://www.kernel.org/doc/html/v4.14/networking/msg_zerocopy.html}{Linux Kernel v4.14, 2017}}. Furthermore, advanced mechanisms like the Data Plane Development Kit (DPDK)\footnote{The Data Plane Development Kit (\href{https://www.dpdk.org}{DPDK}) is an open-source project managed by Linux Foundation} offers a more direct way to bypass the kernel network stack \citep{gregg2014systems}. To fully leverage these mechanisms, a flat memory representation is crucial.

\clearpage
\section{Capabilities of BurTorch}
\label{app:cap-burtorch}

\subsection{High-level constructions}
\label{app:cap-burtorch-high}

At a higher level of abstraction, \libname{BurTorch} introduces fundamental components such as Neurons, Linear Layers, and Multi-Layer Perceptron Layers. These components encapsulate essential operations like linear (or affine) transformations and include best-practice layer parameter initialization. They also support standard activation functions, such as Sigmoid, ReLU, Tanh, or identity mapping, making it easy to model complex computations.

\subsection{Low-level constructions}
\label{app:cap-burtorch-low}

\libname{BurTorch} at the most granular level operates on simple \textit{scalars} from $\mathbb{R}$. Every scalar is indexed sequentially, maintaining simplicity and efficiency throughout the implementation. Unlike other frameworks, \libname{BurTorch} avoids the use of complex fused operations or heavy reliance on Single-Instruction Multiple-Data (SIMD) CPU registers for layer-specific optimizations or custom computation function implementation for specific cases. While SIMD can offer performance gains, its use often leads to unnecessarily complex, unmanageable code. 

Instead, \libname{BurTorch} adopts a streamlined approach, leveraging Instruction-Level Parallelism (ILP) through straightforward loop unrolling and efficient utilization of modern CPU pipelines. In architectures where floating-point functional units can handle both SIMD and ILP effectively, \libname{BurTorch} fully capitalizes on available resources without adding unnecessary implementation complexity.

\subsection{Supported scalars}
\label{app:burotrch-scalars}

\libname{BurTorch} supports the processing of computation graphs that involve operations on \textit{scalars}. Scalars, in this context, can be one of the following types:

\begin{enumerate}
\item \textbf{FP32, FP64:} These represent standard floating-point arithmetic formats, offering approximate real-number computations with single (FP32) and double (FP64) precision, as defined by the IEEE 754-2008 \citep{IEEE754-2008}.

\item \textbf{{SIMD Vector Registers}}: Operations, as listed in Table~\ref{tab:optype}, can be performed on small SIMD registers, typically 128, 256, or 512 bits in length. \libname{BurTorch} does not merely exploit SIMD registers for efficiency but can directly construct computations using these registers. Supported SIMD architectures include SSE2 (128-bit vector registers), AVX2 (256-bit vector registers), AVX-512 (512-bit vector registers), and ARM Neon (128-bit vector registers).

\item \textbf{FP16, BF16, FP128 with Switching to C++23:} While FP32 and FP64 remain the most commonly used formats, they are not the only ones available today. Modern computer hardware capable of processing 16-bit floating-point data is becoming more widely accessible. \libname{BurTorch} can be used with the following extended floating-point types from IEEE 754-2008: Half-precision floating-point format (FP16), and Quadruple-precision floating-point format (FP128). Also \libname{BurTorch} can be used with computing in brain floating-point format (BF16) \citep{wang2019bfloat16}.

\libname{BurTorch} can be compiled using any C++20 and C++23 compatible compiler. However, to use these formats, which were introduced in the C++23 standard, \libname{BurTorch} must be configured to use C++23, and a compiler that supports these formats, such as GCC 13.1\footnote{See \href{https://gcc.gnu.org/releases.html}{https://gcc.gnu.org/releases.html},
\href{https://en.cppreference.com/w/cpp/compiler_support/23}{https://en.cppreference.com/w/cpp/compiler\_support/23}}, must be used.

\end{enumerate}

\subsection{Supported core operators at scalar level}
\label{app:burotrch-scalar-operators}

\begin{table}[h!]
\footnotesize
\centering
\caption{Supported core operations in \libname{BurTorch} at the granularity of single scalar-level computations. \textbf{Args.} = Arguments. For mnemonics see Appendix~\ref{app:burotrch-scalar-operators}.}
\begin{tabular}{@{}lllll@{}}
	\toprule
	\textbf{Mnemonics} & \textbf{Args.} & \textbf{Internal Name}  & \textbf{Description} \\ 
	\midrule
	\texttt{leaf} & \texttt{[s]} & \texttt{eLeaf} & A basic node with input \\ 
	\texttt{relu} & \texttt{[s]} & \texttt{eRelu} & Applies the $\max(0,x)$  \\
	\texttt{tanh} & \texttt{[s]} & \texttt{eTanh} & Applies the hyperbolic tangent \\ 
	\texttt{exp} & \texttt{[s]} & \texttt{eExp} & Applies the computation $\exp(x)$ \\ 
	\texttt{negativeLog} & \texttt{[s]} & \texttt{eNegLog} & The minus natural logarithm \\ 
	\texttt{sigmoid} & \texttt{[s]} & \texttt{eSigmoid} & Applies the sigmoid function  \\ 
	\texttt{inv} & \texttt{[s]} & \texttt{eInv} & Computes $\nicefrac{1}{X}$ \\ 
	\texttt{sqr} & \texttt{[s]} & \texttt{eSqr} & Computes $x^2$ \\ 
	\texttt{pow3} & \texttt{[s]} & \texttt{eCub} & Computes $x^3$ \\ 
	\texttt{logarithm} & \texttt{[s]} & \texttt{eLog} & The natural logarithm of input \\ 
	\texttt{sqrt} & \texttt{[s]} & \texttt{eSqrt} & Computes the square root $\sqrt{x}$ \\ 
	\texttt{invSqrt} & \texttt{[s]} & \texttt{eInvSqrt} & Computes  $\nicefrac{1}{\sqrt{x}}$ \\ 
	\texttt{+, add} & \texttt{[bin]} & \texttt{eBinaryAdd} & Computes for $x,y$ result $x+y$ \\ 
	\texttt{-, sub} & \texttt{[bin]} & \texttt{eBinarySub} & Computes for $x,y$ result $x-y$ \\ 
	\texttt{*, mul} & \texttt{[bin]} & \texttt{eBinaryMult} & Computes for $x,y$ result $x \times y$ \\ 
	\texttt{mulByConstant(x,c)} & \texttt{[bin]} & \texttt{eBinaryMultByConst} & Computes $x \times c$ for constant $c$ \\
	\texttt{/, div} & \texttt{[bin]} & \texttt{eBinaryDiv} & Divides two inputs $x,y$ as $x/y$ \\ 
	\texttt{mean} & \texttt{[bin]} & \texttt{eBinaryMean} & Computes the mean $\nicefrac{(x+y)}{2}$ \\   
	\texttt{addSquares} & \texttt{[bin]} & \texttt{eBinaryAddSquares} & Computes $x^2+y^2$ \\
	\texttt{meanSquares} & \texttt{[bin]} & \texttt{eBinaryMeanSquares} & Computes $\nicefrac{(x^2+y^2)}{2}$ \\ 
	\texttt{negativeMean} & \texttt{[bin]} & \texttt{eBinaryNegativeMean} & Computes $\nicefrac{-(x+y)}{2}$ \\   
	
	\texttt{reduceSum} & \texttt{[var]} & \texttt{eAddVarying} & Computes $\sum_{i=1}^{n} x_i$ \\ 
	\texttt{reduceSub} & \texttt{[var]} & \texttt{eSubVarying} & Computes $x_1 - \sum_{i=2}^{n} x_i$ \\
	\texttt{reduceMul} & \texttt{[var]} & \texttt{eMulVarying} & Computes $\prod_{i=1}^{n} x_i$ \\
	\texttt{reduceMean} & \texttt{[var]} & \texttt{eMeanVarying} & Computes $\nicefrac{1}{n} \sum_{i=1}^{n} x_i$ \\ 
	\texttt{reduceSumOfSquares} & \texttt{[var]} & \texttt{eSumOfSquaresVarying} & Computes $ \sum_{i=1}^{n} x_i^2$ \\
	\texttt{reduceMeanSquares} & \texttt{[var]} & \texttt{eMeanSquaresVarying} & Computes $ \nicefrac{1}{n}\sum_{i=1}^{n} x_i^2$ \\
	\texttt{reduceNegativeMean} & \texttt{[var]} & \texttt{eNegativeMeanVarying} & Computes $\nicefrac{-1}{n} \sum_{i=1}^{n} x_i$ \\ 
	\texttt{innerProduct} & \texttt{[var]} & \texttt{eInnerProductNoBias} & The dot product $\langle x, y \rangle$ \\ 
	\texttt{innerProductWithBias} & \texttt{[var]} & \texttt{eInnerProductWithBias} & Computes $\langle x, y \rangle + b$ \\             
	\bottomrule
	
\end{tabular}
\label{tab:optype}
\end{table}

The core computations in Table~\ref{tab:optype} are atomic operations used to construct the computational graph from simple scalars. This constructed graph, in eager mode, is suitable for both function evaluation and exact gradient computation via Automatic Differentiation. The notations \texttt{[var]}, \texttt{[bin]}, and \texttt{[s]} represent the number of arguments required for each operation:

\begin{enumerate} 
\item \texttt{[var]}: The operation accepts an arbitrary number of arguments. 
\item \texttt{[bin]}: The operation involves exactly two arguments (binary operation). 
\item \texttt{[s]}: The operation is performed on a single argument (unary operation). 
\end{enumerate}

These core computations, as outlined in Table~\ref{tab:optype}, are implemented using standard C++ constructs. All computation code for operands is written as C++ template header-only code. This design choice ensures that the code is both efficient and capable of effective dispatching. The operations are \textit{atomic} in the sense that they are implemented independently, not in the sense of atomic access for operands.

\subsection{Supported derived operators at the scalar level}
\label{app:burotrch-derived-operators}

\libname{BurTorch} supports derived operators at the scalar level, as presented in Tables \ref{tab:inplace-ops} and \ref{tab:heps-compute-ops}. The implementation of these operators requires the allocation of more than one computation node from the pool of scalar nodes. While this aspect can be disregarded from a usage perspective, it is important to note that, practically, some of the information necessary for implementing these operators may already be available within the surrounding context of the expression.

\begin{table}[h!]
\footnotesize
\centering
\caption{In-place operations supported in \libname{BurTorch} at the scalar level.}
\begin{tabular}{@{}lllll@{}}
	\toprule
	\textbf{Mnemonics} & \textbf{Arguments}  & \textbf{Description} \\ 
	\midrule
	\texttt{+=, addInplace} & \texttt{[bin]} & Adds a value in-place, i.e., $x \leftarrow x + y$ \\ 
	\texttt{-=, subInplace} & \texttt{[bin]} & Subtracts a value in-place, i.e., $x \leftarrow x - y$ \\ 
	\texttt{*=, multInplace} & \texttt{[bin]} & Multiplies a value in-place, i.e., $x \leftarrow x \times y$ \\ 
	\texttt{/=, divInplace} & \texttt{[bin]} & Divide a value in-place, i.e., $x \leftarrow \nicefrac{x}{y}$ \\ 
	\bottomrule
\end{tabular}
\label{tab:inplace-ops}
\end{table}

\begin{table}[h!]
\footnotesize
\centering
\caption{Help not-atomic scalar compute operations supported in \libname{BurTorch}.}
\begin{tabular}{@{}lllll@{}}
	\toprule
	\textbf{Mnemonics} & \textbf{Arguments}  & \textbf{Description} \\ 
	\midrule
	\texttt{varianceBiased} & \texttt{[var]} & $\sum_{i=1}^{n} \dfrac{x_i^2}{n} - \left(\sum_{j=1}^{n} \dfrac{x_j}{n} \right)^2$ \\
	\texttt{variance} & \texttt{[var]} & $\dfrac{n}{n-1} \cdot \left(\sum_{i=1}^{n} \dfrac{x_i^2}{n} - \left(\sum_{j=1}^{n} \dfrac{x_j}{n} \right)^2 \right)$ \\
	\texttt{reduceMeanAndMeanSquares} & \texttt{[var]} & $\nicefrac{1}{n}\sum_{i=1}^{n} x_i$ and  $ \nicefrac{1}{n}\sum_{i=1}^{n} x_i^2$ \\
	\bottomrule
\end{tabular}
\label{tab:heps-compute-ops}
\end{table}

\subsection{Supported Matplotlib scripts generation}
\label{app:matplotlib}

\begin{enumerate}
\item \texttt{generateHeatMapBasic:} This method generates a Python script as a string that uses Matplotlib to plot a heatmap.

\item \texttt{generateHeatMap:} This method generates a Python script as a string that uses Matplotlib to plot a heatmap with customized cell values, labels, and counters, applying specific text annotations at each cell based on the itemGetter and counterGetter functions for each matrix element.

\item \texttt{generatePlot:} This method generates a Python script as a string that plots a mathematical function over a specified range (xStart to xEnd) using Matplotlib, with the function's values computed at regular intervals and displayed with a grid and title.

\item \texttt{buildDotGraph:} This method generates a dot graph representation of a tree structure, where each node is described with relevant information, such as label, help, derivative, data references, and connections between nodes.

\item \texttt{asString:} Creates a string representation of a compute node.
\end{enumerate}

\subsection{High-level and low-level connections in BurTorch}
\label{app:cap-burtorch-high-with-low}

\paragraph{Transparency in buffer storage organization.}

At the low level, \libname{BurTorch} ensures complete transparency in the management of partial derivatives and activations, organizing them sequentially in virtual memory for optimal read/write access. The high-level constructs in \libname{BurTorch} are intentionally designed to rely exclusively on low-level implementations, promoting clarity, modularity, and adaptability across the framework. Except the optimized backpropagation internals, every aspect of \libname{BurTorch} remains highly customizable due to its compact and efficient codebase. \libname{BurTorch} is built on a minimalist philosophy, which reduces runtime complexity, dispatch overhead, and overall implementation intricacy. 
Current DL frameworks are heavily optimized for throughput and large batch sizes but are often constrained by APIs shaped by language limitations, particularly in Python. Python’s inefficiency in managing function calls and loops forces users to write mathematical computations in ways that align with the framework rather than the problem itself. This artificial constraint has become the norm. In contrast, \libname{BurTorch} addresses these inefficiencies by combining simplicity, flexibility, and computational efficiency, enabling researchers to explore straightforward models and algorithms without being hindered by complex frameworks, inefficient language ecosystems, or unnecessary abstractions.

\paragraph{Backpropagation with total memory control.}
If you need fine-grained memory control during backpropagation, use the \texttt{backwardWithScratchStorage} function. This C++ template function is designed to perform backpropagation in a computation graph, utilizing scratch storage for intermediate data. It accepts several parameters, including the root node of the graph, reverse topological order data, leaf nodes that can be computed in any order, and a recursion stack.

The \texttt{backwardWithScratchStorage} function optimizes the backpropagation process by leveraging scratch storage and efficient memory management. It ensures that the computation graph is traversed in reverse topological order, processing both internal and leaf nodes appropriately. The recursion stack enables efficient handling of nodes with multiple children. Additionally, the function relies \textit{exclusively} on scratch storage for intermediate data, guaranteeing that memory is allocated and deallocated efficiently during traversal. The function utilizes a recursion stack to maintain the state of the traversal. As each node is processed, it is pushed onto the stack. The stack dynamically grows and shrinks as nodes are processed, ensuring that the reverse topological order is followed and each node is handled in the correct sequence. This function focuses on traversing the graph in topological order, marking the nodes as processed, and managing recursive calculations on the graph nodes. It optimizes memory management by utilizing scratch storage for nodes that need to be processed in a specific order while handling leaf nodes separately, as they can be processed in any order due to the absence of children.

\paragraph{Backpropagation with simple backward.}
When using \libname{BurTorch} for early prototyping, you may prefer not to focus on memory storage concerns. If you don't need to optimize memory usage at this stage, you can invoke the simple backward function. This is standard backpropagation, and while it doesn't include the optimizations for memory allocation, recursive traversal, and efficient node processing provided by \texttt{backwardWithScratchStorage}, it may be sufficient.

\paragraph{\color{black}Saving and loading computation graph values and gradients.} \libname{BurTorch} provides efficient mechanisms for saving and loading elements of the computation graph to and from files and memory. Scalar values are indexed incrementally until an explicit collection of unused indices is invoked, or the construction of new computation nodes is reversed. Each scalar tensor is thus associated with a unique index. The framework allows for saving or loading all tensors within a specified index range (from first to last) to/from a file, or for saving and loading the entire computation graph. This approach offers compile-time flexibility in specifying which elements should be saved or loaded. These operations are highly efficient, as the memory layout for the range between two indices (retrieved using \texttt{sysGetRawNodeIndex}) is sequential.

\subsection{Comparison of {BurTorch} with explicit code snippets}
\label{app:exp2-listing}

\libname{BurTorch} provides an API that closely resembles those of \citet{karpathy2020micrograd} and \libname{PyTorch} \citep{paszke2019pytorch}. The exact code snippets for the three fully functional compute graphs (excluding the setup environment) are presented in Listing~\ref{lst:exp2-listing} below, illustrating their near-identical structure. This demonstrates that leveraging modern C++ can offer researchers a practical and efficient solution without introducing undue complexity.

\begin{figure}[h!]
\centering
\begin{tabular}{ccc}
Micrograd & BurTorch & PyTorch\\
\begin{minipage}{0.32\textwidth}
\tiny
\lstset{basicstyle=\ttfamily\tiny, breaklines=true}
\begin{lstlisting}[language=Python,caption={}]
#!/usr/bin/env python3
# Python code
from micrograd.engine import Value

if __name__ == "__main__":
  a = Value(-4.0)
  b = Value(2.0)
  c = a + b
  d = a * b + b**3
  c += c + 1
  c += 1 + c + (-a)
  d += d * 2 + (b+a).relu()
  d += 3 * d + (b-a).relu()
  e = c - d
  f = e**2
  g = f / 2.0
  g += 10.0 / f
  g.backward()
  print(a.grad)
\end{lstlisting}
\end{minipage}
&
\begin{minipage}{0.32\textwidth}
\tiny
\lstset{basicstyle=\ttfamily\tiny, breaklines=true}
\begin{lstlisting}[language=C++,caption={}]
// C++ code
#include "burtorch.h"
#include <iostream>

int main() {
  auto a = Value(-4.0);
  auto b = Value(2.0);
  auto c = a + b;
  auto d = a * b + pow3(b);
  c += c + Value(1.0);
  c += Value(1.0) + c - a;
  d += d * Value(2.0) + relu(b+a);
  d += Value(3.0) * d + relu(b-a);
  auto e = c - d;
  auto f = sqr(e);
  auto g = f / Value(2.0);
  g += Value(10.0) / f;
  backward(g);
  std::cout<<a.gradCopy();
  return 0;
}
\end{lstlisting}
\end{minipage}
&
\begin{minipage}{0.32\textwidth}	
\tiny
\lstset{basicstyle=\ttfamily\tiny, breaklines=true}	
\begin{lstlisting}[language=Python,caption={}]
#!/usr/bin/env python3
# Python code
import torch

if __name__ == "__main__":
  a = torch.tensor(-4.0, requires_grad=True, dtype=torch.float64)
  b = torch.tensor(2.0, requires_grad=True,dtype=torch.float64)
  c = a + b
  d = a * b + b**3
  c += c + 1
  c += 1 + c + (-a)
  d += d * 2 + (b+a).relu()
  d += 3 * d + (b-a).relu()
  e = c - d
  f = e**2
  g = f / 2.0
  g += 10.0 / f
  g.backward()
  print(a.grad.item())
\end{lstlisting}
\end{minipage}
\end{tabular}

\caption{Listings for the small compute graph shown in Figure~\ref{fig:exp2-small-compute-graph}, adapted from \citet{karpathy2020micrograd}.}

\label{lst:exp2-listing}
\end{figure}

\subsection{Unique aspects of BurTorch}
\label{app:burotrch-unique}

\paragraph{\color{black}{1. Computing gradient without evaluating function value.}}

\libname{BurTorch} provides an imperative approach to constructing scalar values in a computation graph, typically culminating in a loss function. However, the final loss value itself is not required for backpropagation. Unlike traditional frameworks, \libname{BurTorch} allows gradients to be computed without explicitly evaluating the function value, providing a minor optimization. While this optimization can impact performance depending on the computation graph’s structure, it was not used in any experiments in this work.  

\paragraph{\color{black}{2. BurTorch compliance with high-reliability industry standards.}}

Motor Industry Software Reliability Association (MISRA) guidelines \citep{misra2012} are widely adopted in safety-critical industries such as automotive, aerospace, and medical devices. Most Deep Learning frameworks fail to meet MISRA’s stringent requirements, particularly its prohibitions on (i) dynamic memory allocation at runtime (Rule 4.12 \citet{misra2012}) and (ii) recursive functions (Rule 17.2 \citet{misra2012}), both of which ensure system stability and predictability. 

\libname{BurTorch} addresses these concerns by eliminating runtime memory allocation—configurable to operate solely on pre-allocated scratch buffers—and by avoiding recursion, ensuring MISRA compliance.

\paragraph{\color{black}{3. Maximizing performance with pre-allocated buffers.}}

Most Deep Learning frameworks rely on dynamic memory allocation for parameters, activations, and gradients. While they offer some control over memory management, they still depend on dynamic allocation through internal logic or external libraries. \libname{BurTorch}, in contrast, exclusively operates on pre-allocated, user-provided buffers, eliminating runtime memory allocation. This design is particularly beneficial in environments with strict memory and communication constraints, such as embedded systems or Federated Learning. By minimizing auxiliary memory usage and avoiding unnecessary copying, \libname{BurTorch} achieves optimal performance in resource-limited settings and ensures compatibility with systems where dynamic memory allocation is prohibited.

\paragraph{\color{black}{4. A small codebase enables easy customization.}}

Due to its compact codebase, \libname{BurTorch} can be modified to address research needs that cannot be easily achieved through modifications in complex frameworks. See Section~\ref{sec:inluence-on-theory}.  

\paragraph{\color{black}{5. {BurTorch} is thread-safe.}}

\libname{BurTorch} can be built with flags to ensure true thread safety, a distinguishing feature that sets it apart from many alternative frameworks. In \libname{BurTorch}, computations can be safely executed in parallel without concerns about race conditions, thanks to its carefully designed architecture and, importantly, its compact codebase. In frameworks such as \libname{TensorFlow}, \libname{PyTorch}, and \libname{JAX}, special consideration must be given to thread safety. These frameworks rely on complex chains of dependencies, where the failure of even a single component to guarantee thread safety compromises the overall framework's ability to provide strong thread safety assurances. As a consequence, the thread safety of \libname{PyTorch}, \libname{TensorFlow}, and \libname{JAX} depends on the specific use case.

\paragraph{\color{black}{6. Concurrent graph construction across multiple threads.}}

\libname{BurTorch} allows multiple real operating system threads to simultaneously contribute to the construction of a single computation graph. By enabling concurrent graph-building, \libname{BurTorch} takes full advantage of modern hardware, leading to a more efficient distribution of computational tasks. This capability to manage parallel graph construction offers a level of flexibility that is often lacking in other frameworks. It is a key advantage, enabling \libname{BurTorch} to scale more effectively in multi-threaded environments, especially when optimizing graph construction is necessary.

\paragraph{\color{black}{7. Ability to eliminate named parameters.}}

\libname{PyTorch} \citep{paszke2019pytorch}, \libname{TensorFlow} \citep{abadi2016tensorflow}, and \libname{JAX} \citep{jax2018github} with the FLAX extension \citep{flax2020github} do not provide a mechanism to remove names from trainable parameters. While named parameters are useful during research and debugging, once the model is finalized, they may no longer be necessary if the optimization algorithms do not rely on names.

\libname{BurTorch} supports names for variables, but it can be configured and built to completely remove the functionality of named parameters or variables at runtime. And it requires no changes to the source code to eliminate this redundant logic.

Although this may appear to be a minor feature if each scalar has a unique text description longer than two characters, the memory consumed by these descriptions exceeds the memory required to store the $x\in\mathbb{R}^d$ in FP16 format.

\clearpage
\section{Missing Experiment on Linux OS}
\label{app:extra-experiments-linux}

In this appendix, we present the results of comparing \libname{BurTorch} against popular Deep Learning frameworks under experimental settings similar to those described in Section \ref{sec:experiments} of the main text, with two differences.

First, the experiments were conducted on a Linux workstation running {Ubuntu 20.04.6}. Although the CPU used is also x86-based, its clock frequency is slightly lower, fixed at $3.2$ GHz during the experiments. We employed the methodology from \citet{burlachenko2024unlocking} to ensure the reproducibility of the measurement during experiments.

Second, for memory consumption, we report the peak virtual memory usage (VmSize). The memory subsystem is closely tied to the operating system, so it may be useful to revisit what constitutes this value. For more detailed explanations, we refer the reader to specialized literature, such as \citet{kerrisk2010linux}.

\paragraph{Background on the VmSize metric.} In POSIX-based OS, VmSize represents the total amount of virtual memory allocated to a process, including:

\begin{enumerate} 
\item \textbf{Program text.} Memory allocated for the executable code of the process in machine-language instructions. This memory may be shared with other applications, except for data within binary applications. In such cases, system mechanisms like copy-on-write may be involved.

\item \textbf{Initialized data.} The segment containing global and static variables that are explicitly initialized.

\item \textbf{Uninitialized data.} Global and static variables that are not explicitly initialized. Before the program starts, the system initializes such memory to zero, explaining why the size of a binary executable file does not directly reflect its runtime memory requirements.

\item \textbf{Heap memory.} Memory allocated for dynamically allocated data.

\item \textbf{Stack memory.} A dynamically growing and shrinking segment that contains stack frames. Each stack frame is allocated for the currently called function and stores local variables, function arguments, and return values.

\item \textbf{Memory-mapped files.} Additional files or resources mapped into the process’s memory space. Mapped files allow their contents to be accessed as bytes in a corresponding memory region. Pages from named memory-mapped files are loaded into memory on demand. Anonymous mappings, which are not associated with files, are backed by the swap.

\end{enumerate}

The peak \texttt{VmSize} reflects the total virtual address space of the process, encompassing both actively used memory and memory that may not be physically resident in DRAM but is backed either by swap files or another file (if the content source is a physical file mapped via the memory-mapping mechanism).

\clearpage

\begin{table}[h!]
\footnotesize
\centering
\caption{Backpropagation over $100$K iterations with a {tiny} dynamic compute graph from Figure~\ref{fig:tiny-compute-graph}. Mean and standard deviation across $5$ launches. Computation in FP64, one CPU Core $3.2$ GHz (x86-64). Physical Memory: DDR4, 251 GBytes. Linux Ubuntu 20.04. See also Figure~\ref{fig:execution_times_speedup_linux}. The numerical results across frameworks match exactly.
}
\begin{tabular}{|l|l|l|l|l|}
	\hline
	\textbf{\#} & \textbf{Framework, Mode, Language} & \textbf{Device} & \textbf{\makecell[c]{Compute Time \\ (sec.)}} & \textbf{\makecell[c]{Relative \\ to \\ BurTorch}} \\ 
	\hline
	\hline
	\cellcolor{bgcolorwe}1&\cellcolor{bgcolorwe}BurTorch, Eager, C++ & \cellcolor{bgcolorwe}CPU                   & \cellcolor{bgcolorwe}$0.011 \pm 0.00007$             & \cellcolor{bgcolorwe}$\times 1.0$ (We)             \\ \hline
	2& TensorFlow 2.18.0, Eager, Python & CPU          & $163.538 \pm 1.584$         & $\times 14\,867.0\,\,\,$                    \\ \hline
	3& TensorFlow 2.18.0, Graph, Semi-Python & CPU          & $27.882 \pm 0.280$         & $\times 2\,534.7\,\,\,\,\,$                    \\ \hline
	4& \makecell[l]{TF Lite 2.18.0, Graph,\\TF Lite Interpreter} & CPU          & $0.408 \pm 0.005$         & $\times 37$                    \\ \hline
	5& Autograd 1.7.0, Eager, Python & CPU            & $29.250 \pm 0.464$        & $\times 2659.09\,\,\,$                    \\ \hline
	6& {PyTorch} 2.5.1, Eager, Python & \textbf{GPU}          & $35.240 \pm 2.193$              & $\times 3\,203.6\,\,\,$                    \\ \hline
	7& {PyTorch} 2.5.1, Eager, Python & CPU          & $15.440 \pm 0.037$              & $\times 1\,403.6\,\,\,$                    \\ \hline
	8& {PyTorch} 2.5.1, Graph, TorchScript & CPU & $14.224 \pm 0.059$                 & $\times 1\,293.0\,\,\,$                    \\ \hline
	9& {PyTorch} 2.5.1, Eager, LibTorch, C++ & CPU & $7.611 \pm 0.079$                 & $\times 691.9\,\,\,$                    \\ \hline
	10& {JAX} 0.4.30, Eager, Python & CPU                & $506.464 \pm 1.555$         & $\times 46\,042.1$                      \\ \hline
	11& {JAX} 0.4.30, Graph, Semi-Python & CPU                & $11.156 \pm 0.0655$         & $\times 1014.1\,\,\,\,\,\,\,$                      \\ \hline
	12& {Micrograd}, Eager, Python & CPU                & $2.751 \pm 0.007$                & $\times 250.0\,\,\,\,\,\,\,$                      \\ \hline
	13& {Apple MLX} 0.22, Eager, Python & CPU                & $3.072 \pm 0.028$                & $\times 279.2\,\,\,\,\,\,\,$                      \\ \hline
\end{tabular}
\label{tab:execution_times_speedup_linux}
\end{table}

\begin{figure*}[ht]
\centering
\includegraphics[width=0.85\linewidth]{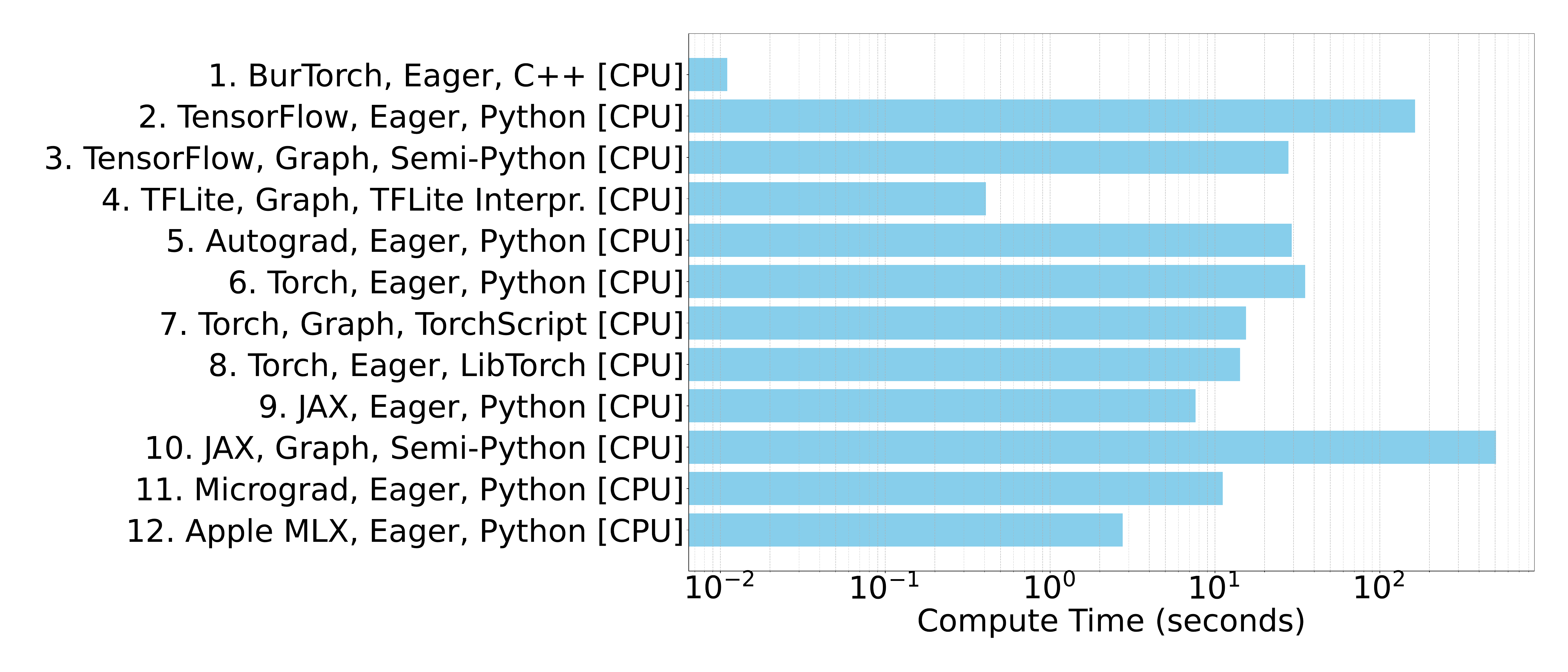}

\caption{Visualization of Table~\ref{tab:execution_times_speedup_linux}. Backpropagation over $100$K iterations with a {tiny} dynamic compute graph from Figure~\ref{fig:tiny-compute-graph}. Computation in FP64, one CPU Core $3.2$ GHz (x86-64). Linux Ubuntu 20.04.}
\label{fig:execution_times_speedup_linux}
\end{figure*}

\subsection{Experiments with tiny compute graph on Linux}
\label{app:extra-experiments-linux-tiny-graph}

This experiment replicates the setup from Section \ref{sec:experiments-tiny}. The results are presented in Table~\ref{tab:execution_times_speedup_linux} and Figure~\ref{fig:execution_times_speedup_linux}. The software and hardware environment used are detailed in Appendix \ref{app:exp-setup-linux}. In addition to the software used in Windows OS experiments, we have included a standalone Automatic Differentiation tool distributed by Apple, named \libname{Apple MLX} \citep{mlx2023}. The results in Table~\ref{tab:execution_times_speedup_linux} show that \libname{BurTorch} achieves significant speedups, thanks to its latency-optimized design.

\begin{table}[h!]
\footnotesize
\centering
\caption{Comparison of \libname{BurTorch} and \libname{PyTorch} performance for training MLP-like model. Batch: $b=1$, Compute: FP32, Single CPU core (Intel(R) Xeon(R) Gold 6146 CPU $3.20$ GHz). Initialization time is the end-to-end time for training a model with $1$ iterations. Compute time excludes batch preparation. Memory represents the peak private virtual memory per training. Linux Ubuntu 20.04.}
\label{tab:exp3-compute-and-mem-speedup-b1-linux}
\begin{tabular}{|l|l|l|l|l|l|l|l|}
	\hline
	\textbf{\#} & \textbf{Parameters (d)} & \multicolumn{3}{c|}{\textbf{\makecell[c]{PyTorch, \\Eager, v2.5.1 [CPU]}}} & \multicolumn{3}{c|}{\cellcolor{bgcolorwe}{\textbf{{BurTorch, Eager [CPU]}}}} \\ 
	\cline{3-8}
	& \textbf{Hidden Dim.(e)} & \textbf{\makecell[c]{Init\\(ms)}} & \textbf{\makecell[c]{Compute\\(ms)}} & \textbf{\makecell[c]{Mem.\\(MB)}} & {\textbf{\makecell[c]{Init\\(ms)}}} & {\textbf{\makecell[c]{Compute\\(ms)}}} & {\textbf{\makecell[c]{Mem.\\(MB)}}} \\ 
	\hline
	\hline
	1&$5,963\ (e=4)$ & $4\,649$ & $1.54 \pm 1.60$ & $13\,586$ & $8$ & $0.075 \pm 0.007$ & $41.8$ \\ 
	2&$18,587\ (e=16)$ & $4\,828$ & $1.54 \pm 1.43$ & $13\,647$ & $8$ & $0.150 \pm 0.001$ & $42.7$ \\ 
	3&$35,419\ (e=32)$ & $4\,469$ & $1.56 \pm 1.49$ & $13\,578$ & $16$ & $0.242 \pm 0.024$ & $44.1$ \\ 
	4&$69,083\ (e=64)$ & $4\,713$ & $1.67 \pm 1.62$ & $13\,643$ & $16$ & $0.378 \pm 0.038$ & $46.6$ \\ 
	5&$136,411\ (e=128)$ & $4\,879$ & $1.88 \pm 1.58$ & $13\,579$ & $27$ & $0.724 \pm 0.082$ & $51.3$ \\ 
	6&$540,379\ (e=512)$ & $4\,768$ & $2.37 \pm 1.47$ & $13\,598$ & $33$ & $2.382 \pm 0.306$ & $79.7$ \\ 
	7&$1,079,003\ (e=1024)$ & $5\,081$ & $3.23 \pm 1.52$ & $13\,661$ & $47$ & $6.192 \pm 0.948$ & $116.5$ \\ 
	\hline
\end{tabular}
\end{table}

\begin{table}[h!]
\footnotesize
\centering
\caption{Comparison of \libname{BurTorch} and \libname{PyTorch} performance for training MLP-like model. Batch: $b=64$, Compute: FP32, Single CPU core (Intel(R) Xeon(R) Gold 6146 CPU $3.20$ GHz). Initialization time is the end-to-end time for training a model with $1$ iterations. Compute time excludes batch preparation. Memory represents the peak private virtual memory. DRAM: 251 GB. Linux Ubuntu 20.04.}
\label{tab:exp3-compute-and-mem-speedup-b64-linux}
\begin{tabular}{|l|l|l|l|l|l|l|l|}
	\hline
	\textbf{\#} & \textbf{Parameters (d)} & \multicolumn{3}{c|}{\textbf{\makecell[c]{PyTorch, \\Eager, v2.5.1 [CPU]}}} & \multicolumn{3}{c|}{\cellcolor{bgcolorwe}{\textbf{{BurTorch, Eager [CPU]}}}} \\ 
	\cline{3-8}
	& \textbf{Hidden Dim.(e)} & \textbf{\makecell[c]{Init\\(ms)}} & \textbf{\makecell[c]{Compute\\(ms)}} & \textbf{\makecell[c]{Mem.\\(MB)}} & {\textbf{\makecell[c]{Init\\(ms)}}} & {\textbf{\makecell[c]{Compute\\(ms)}}} & {\textbf{\makecell[c]{Mem.\\(MB)}}} \\ 
	\hline
	\hline
	1& $5,963\ (e=4)$ & $5\,658$ & $9.98 \pm 1.68$ & $13\,602$ & $11$ & $0.976 \pm 0.023$ & $41.8$ \\
	2& $18,587\ (e=16)$ & $5\,774$ & $9.98 \pm 1.96$ & $13\,664$ & $13$ & $3.353 \pm 0.076$ & $42.7$ \\
	3& $35,419\ (e=32)$ & $5\,941$ & $10.37 \pm 2.02$ & $13\,603$ & $21$ & $6.566 \pm 0.156$ & $44.1$ \\
	4& $69,083\ (e=64)$ & $6\,020$ & $10.25 \pm 2.14$ & $13\,605$ & $28$ & $10.546 \pm 0.283$ & $46.6$ \\
	5& $136,411\ (e=128)$ & $6\,176$ & $22.49 \pm 3.55$ & $13\,625$ & $43$ & $21.557 \pm 0.496$ & $51.3$ \\
	6& $540,379\ (e=512)$ & $6\,151$ & $117.13 \pm 6.41$ & $13\,734$ & $136$ & $111.846 \pm 2.566$ & $79.7$ \\ 
	7&$1,079,003\ (e=1024)$ & $5\,926$ & $236.10 \pm 12.76$ & $13\,889$ & $272$ & $240.885 \pm 6.081$ & $116.5$ \\ 
	\hline
\end{tabular}
\end{table}

\subsection{Experiments with medium compute graph on Linux}
\label{app:extra-experiments-linux-mlp-llm}

In this experiment, we evaluated \libname{BurTorch} on a character-level autoregressive prediction model, designed as a medium-complexity compute graph, based on the architecture in \citet{bengio2000neural}, similar to the experiment in Section~\ref{sec:exp3-medium-compute-graphs}. The results, presented in Tables \ref{tab:exp3-compute-and-mem-speedup-b1-linux} and \ref{tab:exp3-compute-and-mem-speedup-b64-linux}, demonstrate \libname{BurTorch}'s efficiency in terms of initialization time and memory usage.

\subsection{Experiments with a GPT-3-like model on Linux}
\label{app:extra-experiments-linux-gpt3-like}

The experiments in this section replicate those from Section~\ref{sec:exp4-burtorch-fot-gpt3}, but were conducted on Linux OS (see Appendix~\ref{app:exp-setup-linux} for setup details). The results, presented in Table~\ref{tab:app-exp4-compute-and-mem-speedup-linux}, confirm that the findings are consistent with those in Section~\ref{sec:exp4-burtorch-fot-gpt3} across all considered OS (macOS, Linux, Windows). \libname{BurTorch} demonstrates strong potential for latency-sensitive applications, and due to its small code size, it may also be valuable for research purposes. As the batch size increases, the performance advantage of \libname{PyTorch} in terms of absolute runtime becomes more pronounced. Nevertheless, \libname{BurTorch} remains significantly more efficient in terms of real memory footprint, achieving approximately $\times 100$ smaller memory requirements.

\begin{table}[h!]
	\footnotesize
	\centering
\caption{Comparison of \libname{BurTorch} and \libname{PyTorch} performance for training \modelname{GPT-3} like model. FP32, Single CPU core (Intel(R) Xeon(R) Gold 6146 CPU $3.20$ GHz). Peak virtual memory per training. DRAM: 251 GB. Linux Ubuntu 20.04. Trainable variables: $46$K.}
	\label{tab:app-exp4-compute-and-mem-speedup-linux}
	\begin{tabular}{|l|cc|cc|cc|}
		\hline
		\textbf{Batch} & \multicolumn{2}{c|}{\cellcolor{bgcolorwe}{\textbf{BurTorch, Eager, C++}}} & \multicolumn{2}{c|}{\textbf{\makecell[c]{PyTorch,\\ Graph, TorchScript}}} & \multicolumn{2}{c|}{\textbf{\makecell[c]{PyTorch,\\ Eager, Python}}} \\ 
		\cline{2-7}
		& {\textbf{\makecell[c]{Compute\\(ms)}}} & {\textbf{\makecell[c]{Mem.\\(MB)}}} & \textbf{\makecell[c]{Compute\\(ms)}} & \textbf{\makecell[c]{Mem.\\(MB)}} & \textbf{\makecell[c]{Compute\\(ms)}} & \textbf{\makecell[c]{Mem.\\(MB)}} \\ 
		\hline
		\hline
		$1$ & $0.913 \pm 0.081$ & $22.3$ & $13.130 \pm 55.560$ & $9\,258$ & $16.144 \pm 0.653$ & $9\,111$ \\ 
		$2$ & $1.605 \pm 0.092$ & $22.3$ & $13.616 \pm 55.606$ & $9\,259$ & $16.570 \pm 0.656$ & $9\,111$ \\ 
		$4$ & $3.222 \pm 0.143$ & $22.3$ & $14.074 \pm 55.552$ & $9\,255$ & $17.034 \pm 0.713$ & $9\,111$ \\ 
		$8$ & $6.110 \pm 0.178$ & $22.3$ & $14.984 \pm 55.631$ & $9\,255$ & $17.842 \pm 0.693$ & $9\,108$ \\
		$16$ & $12.426 \pm 0.253$ & $22.3$ & $16.602 \pm 55.406$ & $9\,253$ & $19.475 \pm 0.701$ & $9\,104$ \\ 
		$32$ & $23.792 \pm 0.404$ & $22.3$ & $19.769 \pm 55.477$ & $9\,255$ & $22.481 \pm 0.724$ & $9\,105$ \\ 
		$64$ & $58.489 \pm 0.151$ & $22.3$ & $26.358 \pm 55.518$ & $9\,259$ & $28.372 \pm 0.801$ & $9\,107$ \\
		\hline
	\end{tabular}
\end{table}

\clearpage
\section{Missing Experiment on macOS}
\label{app:extra-experiments-macos}

In this appendix, we present the results of comparing \libname{BurTorch} against \libname{PyTorch} under experimental settings similar to those described in the main text (Section \ref{sec:experiments}), but conducted on a macOS workstation running {macOS Sonoma 14.5}.

The CPU used in these experiments is x86-based, with a clock frequency of $2.3$ GHz during testing. For a detailed description of the hardware and software environment, please refer to Appendix~\ref{app:exp-setup-macos}.

For memory consumption, we report the peak resident memory usage (RES). RES represents the total amount of physical memory allocated to the process. The peak \texttt{VmSize} value on macOS can be somewhat ambiguous, as Apple produces a tightly integrated software-hardware ecosystem, making the exact meaning of \texttt{VmSize} less straightforward.

\subsection{Experiments with tiny compute graph on macOS}
\label{app:extra-experiments-macos-tiny-graph}

\begin{table}[h!]
\footnotesize
\centering
\caption{Backpropagation over $100$K iterations with a {tiny} dynamic compute graph from Figure~\ref{fig:tiny-compute-graph}. Mean and standard deviation across $5$ launches. Computation in FP64, one CPU Core $2.3$ GHz (x86-64). Physical Memory: 32 GB. macOS Sonoma 14.5. The numerical results across frameworks match exactly. See also Figure~\ref{fig:execution_times_speedup_macos}.
}
\begin{tabular}{|l|l|l|l|l|}
	\hline
	\textbf{\#} & \textbf{Framework, Mode, Language} & \textbf{Device} & \textbf{\makecell[c]{Compute Time \\ (sec.)}} & \textbf{\makecell[c]{Relative \\ to \\ BurTorch}} \\ 
	\hline
	\hline
	\cellcolor{bgcolorwe}1.&\cellcolor{bgcolorwe}BurTorch, Eager, C++ & \cellcolor{bgcolorwe}CPU                   & \cellcolor{bgcolorwe}$0.0118 \pm 0.00024$             & \cellcolor{bgcolorwe}$\times 1.0$ (We)             \\ 
	\hline
	2&TensorFlow 2.16.2, Eager, Python & CPU          & $145.312 \pm 2.287$         & $\times 12\,314.576\,\,\,$                    \\ \hline
	3&\makecell[l]{TensorFlow 2.16.2, Graph, \\Semi-Python} & CPU          & $33.041 \pm 0.386$         & $\times 2\,800.0847\,\,\,\,\,$                    \\ \hline
	4&\makecell[l]{TF Lite 2.16.2, Graph,\\TF Lite Interpreter} & CPU          & $0.728 \pm 0.0111$         & $\times 65$                    \\ \hline
	
	5&Autograd 1.7.0, Eager, Python & CPU            & $30.193 \pm 0.333$        & $\times 2\,558.728\,\,\,$                    \\ \hline
	6&{PyTorch} 2.2.2, Eager, Python & CPU          & $8.712 \pm 0.068$              & $\times 738.305\,\,\,$                    \\ \hline
	7&{PyTorch} 2.2.2, Graph, TorchScript & CPU & $4.978 \pm 0.050$                 & $\times 421.86\,\,\,$                    \\ \hline
	8&{PyTorch} 2.5.1, Eager, LibTorch, C++ & CPU & $5.439 \pm 0.127$                 & $\times 460.932\,\,\,$                    \\ \hline
	9&{JAX} 0.4.30, Eager, Python & CPU                & $445.015 \pm 1.921$         & $\times 37\,713.135$                      \\ \hline
	10&{JAX} 0.4.30, Graph, Semi-Python & CPU                & $12.091 \pm 0.248$         & $\times 1024.66\,\,\,\,\,\,\,$                      \\ \hline
	11&{Micrograd}, Eager, Python & CPU                & $2.399 \pm 0.039$                & $\times 203.305\,\,\,\,\,\,\,$                      \\ \hline
	12&{Apple MLX} 0.7, Eager, Python & CPU                & $3.138 \pm 0.0245$                & $\times 281.186\,\,\,\,\,\,\,$                      \\ \hline
\end{tabular}
\label{tab:execution_times_speedup_macos}
\end{table}

\begin{figure*}[ht]
\centering
\includegraphics[width=0.85\linewidth]{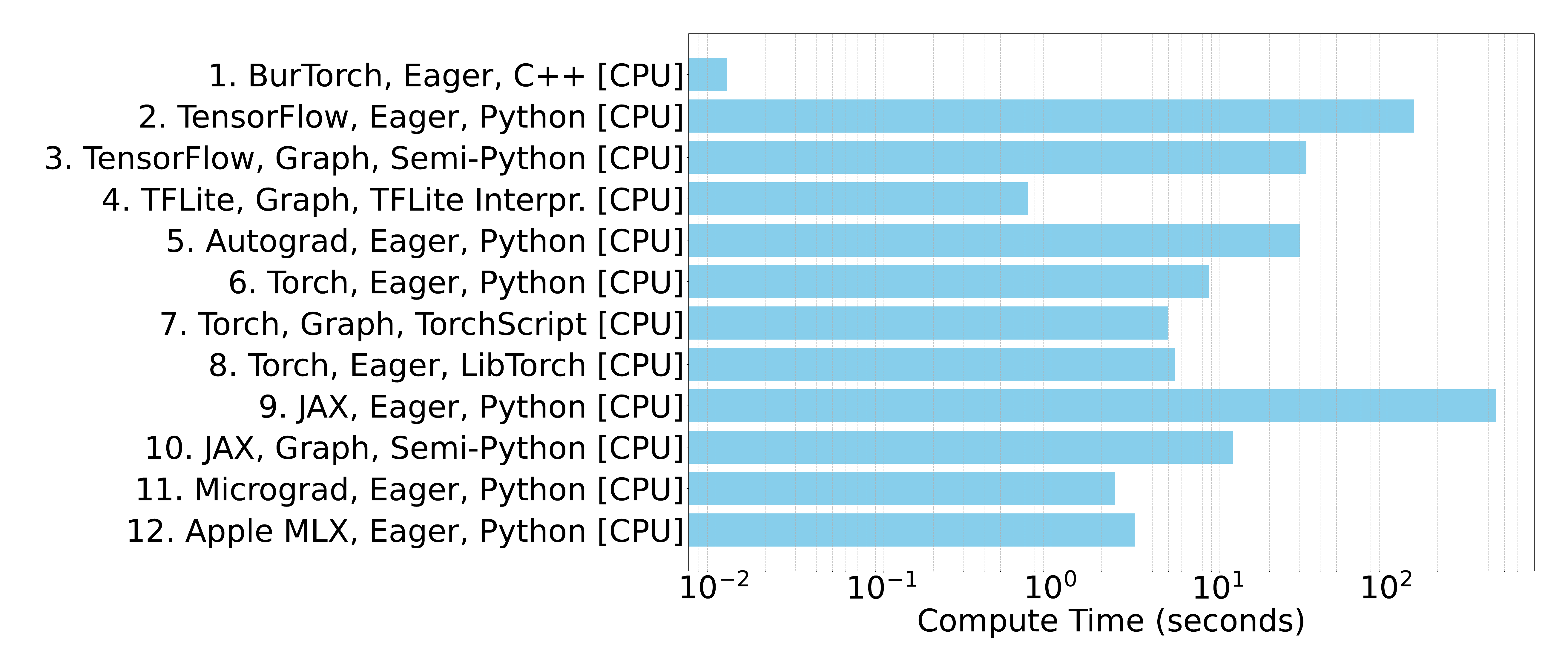}

\caption{Visualization of Table~\ref{tab:execution_times_speedup_macos}. Backpropagation over $100$K iterations with a {tiny} compute graph from Figure~\ref{fig:tiny-compute-graph}. Computation in FP64, one CPU Core $2.3$ GHz (x86-64), macOS Sonoma 14.5. The numerical results across frameworks match exactly.}
\label{fig:execution_times_speedup_macos}
\end{figure*}

This experiment replicates the setup from Section~\ref{sec:experiments-tiny}. The results are presented in Table~\ref{tab:execution_times_speedup_macos} and Figure~\ref{fig:execution_times_speedup_macos}. The software and hardware environment used is detailed in Appendix \ref{app:exp-setup-macos}. In addition to the software used in Windows OS experiments, we have included \libname{Apple MLX} \citep{mlx2023} for comparison, as done in Linux OS experiment.

\subsection{Experiments with medium compute graph on macOS}
\label{app:extra-experiments-macos-mlp-llm}

In this experiment, we evaluated \libname{BurTorch} on a character-level autoregressive prediction model, designed as a medium-complexity compute graph from \citet{bengio2000neural}, similar to the experiment in Section~\ref{sec:exp3-medium-compute-graphs}. The results, presented in Tables \ref{tab:exp3-compute-and-mem-speedup-b1-mac} and \ref{tab:exp3-compute-and-mem-speedup-b64-mac}, demonstrate \libname{BurTorch}'s efficiency in terms of initialization time and memory usage.

\begin{table}[h!]
\footnotesize
\centering
\caption{Comparison of \libname{BurTorch} and \libname{PyTorch} performance for training MLP like model. Batch: $b=1$, Compute: FP32, Single CPU core (Intel Quad-Core Intel Core i7 $2.3$ GHz). Initialization time is the end-to-end time for training a model with $1$ iterations. Compute time excludes batch preparation. Memory represents the peak private virtual memory per training. macOS Sonoma 14.5.}
\label{tab:exp3-compute-and-mem-speedup-b1-mac}
\begin{tabular}{|l|l|l|l|l|l|l|l|}
	\hline
	\textbf{\#} & \textbf{Parameters (d)} & \multicolumn{3}{c|}{\textbf{\makecell[c]{PyTorch, \\Eager, v2.5.1 [CPU]}}} & \multicolumn{3}{c|}{\cellcolor{bgcolorwe}{\textbf{{BurTorch, Eager [CPU]}}}} \\ 
	\cline{3-8}
	& \textbf{Hidden Dim.(e)} & \textbf{\makecell[c]{Init\\(ms)}} & \textbf{\makecell[c]{Compute\\(ms)}} & \textbf{\makecell[c]{Mem.\\(MB)}} & {\textbf{\makecell[c]{Init\\(ms)}}} & {\textbf{\makecell[c]{Compute\\(ms)}}} & {\textbf{\makecell[c]{Mem.\\(MB)}}} \\ 
	\hline
	\hline
	1& $5,963\ (e=4)$ & $6\,424$ & $1.08 \pm 0.35$ & $554$ & $15$ & $0.041 \pm 0.011$ & $36.2$ \\ 
	2& $18,587\ (e=16)$ & $6\,523$ & $1.08 \pm 0.36$ & $557$ & $15$ & $0.059 \pm 0.018$ & $36.8$ \\ 
	3& $35,419\ (e=32)$ & $6\,482$ & $1.12 \pm 0.38$ & $560$ & $16$ & $0.073 \pm 0.020$ & $38.2$ \\ 
	4& $69,083\ (e=64)$ & $6\,381$ & $1.23 \pm 0.37$ & $558$ & $18$ & $0.115 \pm 0.0402$ & $41.8$ \\ 
	5& $136,411\ (e=128)$ & $6\,451$ & $1.39 \pm 0.39$ & $558$ & $19$ & $0.190 \pm 0.059$ & $48.3$ \\ 
	6& $540,379\ (e=512)$ & $6\,443$ & $2.62 \pm 0.67$ & $579$ & $36$ & $0.969 \pm 0.200$ & $87.9$ \\ 
	7& $1,079,003\ (e=1024)$ & $6\,387$ & $4.40 \pm 0.88$ & $612$ & $61$ & $1.995 \pm 0.291$ & $132.0$ \\ 
	\hline
\end{tabular}
\end{table}

\begin{table}[h!]
\footnotesize
\centering
\caption{Comparison of \libname{BurTorch} and \libname{PyTorch} performance for training MLP-like model. Batch: $b=64$, Compute: FP32, Single CPU core (Intel Quad-Core Intel Core i7 $2.3$ GHz). Initialization time is the end-to-end time for training a model with $1$ iterations. Compute time excludes batch preparation. Memory represents the peak private virtual memory. DRAM: 32 GB. macOS Sonoma 14.5.}
\label{tab:exp3-compute-and-mem-speedup-b64-mac}
\begin{tabular}{|l|l|l|l|l|l|l|l|}
	\hline
	\textbf{\#} & \textbf{Parameters (d)} & \multicolumn{3}{c|}{\textbf{\makecell[c]{PyTorch, \\Eager, v2.5.1 [CPU]}}} & \multicolumn{3}{c|}{\cellcolor{bgcolorwe}{\textbf{{BurTorch, Eager [CPU]}}}} \\ 
	\cline{3-8}
	& \textbf{Hidden Dim.(e)} & \textbf{\makecell[c]{Init\\(ms)}} & \textbf{\makecell[c]{Compute\\(ms)}} & \textbf{\makecell[c]{Mem.\\(MB)}} & {\textbf{\makecell[c]{Init\\(ms)}}} & {\textbf{\makecell[c]{Compute\\(ms)}}} & {\textbf{\makecell[c]{Mem.\\(MB)}}} \\ 
	\hline
	\hline
	1& $5,963\ (e=4)$ & $6\,416$ & $7.87 \pm 1.18$ & $656$ & $16$ & $0.300 \pm 0.043$ & $36.2$ \\
	2& $18,587\ (e=16)$ & $6\,429$ & $8.21 \pm 1.04$ & $658$ & $16$ & $0.884 \pm 0.097$ & $37.3$ \\
	3& $35,419\ (e=32)$ & $6\,447$ & $9.24 \pm 1.22$ & $659$ & $18$ & $1.751 \pm 0.16$ & $38.9$ \\
	4& $69,083\ (e=64)$ & $6\,504$ & $11.43 \pm 1.08$ & $659$ & $20$ & $3.512 \pm 0.293$ & $41.7$ \\
	5& $136,411\ (e=128)$ & $6\,488$ & $16.63 \pm 1.35$ & $672$ & $29$ & $6.756 \pm 0.447$ & $47.4$ \\
	6& $540,379\ (e=512)$ & $6\,452$ & $43.02 \pm 1.66$ & $689$ & $75$ & $35.049 \pm 2.439$ & $102.3$ \\ 
	7& $1,079,003\ (e=1024)$ & $6\,512$ & $80.64 \pm 5.93$ & $715$ & $144$ & $80.363 \pm 2.885$ & $140.0$ \\ 
	\hline
\end{tabular}
\end{table}

\clearpage
\subsection{Experiments with a GPT-3-like model on macOS}
\label{app:extra-experiments-macos-gpt3-like}

The experiments in this section replicate those from Section~\ref{sec:exp4-burtorch-fot-gpt3}, but were conducted on macOS (see Appendix~\ref{app:exp-setup-macos} for setup details). The results, presented in Table~\ref{tab:app-exp4-compute-and-mem-speedup-mac}, show that the findings are consistent with those in Section~\ref{sec:exp4-burtorch-fot-gpt3}. \libname{BurTorch} demonstrates strong potential for latency-sensitive applications, and due to its small code size, it may also be valuable for research purposes. As the batch size increases, the performance advantage of \libname{PyTorch} in terms of absolute runtime becomes more noticeable. Nevertheless, as seen in Windows and Linux experiments, \libname{BurTorch} on macOS remains significantly more efficient in terms of real memory footprint.

\begin{table}[h!]
	\footnotesize
	\centering
\caption{Comparison of \libname{BurTorch} and \libname{PyTorch} performance for training \modelname{GPT-3} like model. FP32, Single CPU core (Intel Quad-Core Intel Core i7 $2.3$ GHz). Peak virtual memory per training. DRAM: 32 GB. macOS Sonoma 14.5. Trainable variables: $46$K.}
	\label{tab:app-exp4-compute-and-mem-speedup-mac}
	\begin{tabular}{|l|cc|cc|cc|}
		\hline
		\textbf{Batch} & \multicolumn{2}{c|}{\cellcolor{bgcolorwe}{\textbf{BurTorch, Eager, C++}}} & \multicolumn{2}{c|}{\textbf{\makecell[c]{PyTorch,\\ Graph, TorchScript}}} & \multicolumn{2}{c|}{\textbf{\makecell[c]{PyTorch,\\ Eager, Python}}} \\ 
		\cline{2-7}
		& {\textbf{\makecell[c]{Compute\\(ms)}}} & {\textbf{\makecell[c]{Mem.\\(MB)}}} & \textbf{\makecell[c]{Compute\\(ms)}} & \textbf{\makecell[c]{Mem.\\(MB)}} & \textbf{\makecell[c]{Compute\\(ms)}} & \textbf{\makecell[c]{Mem.\\(MB)}} \\ 
		\hline
		\hline
		$1$ & $0.739 \pm 0.157$ & $17.5$ & $14.598 \pm 61.975$ & $513$ & $12.765 \pm 2.396$ & $235$ \\ 
		$2$ & $1.429 \pm 0.227$ & $18.2$ & $13.363 \pm 53.726$ & $508$ & $13.147 \pm 2.412$ & $242$ \\ 
		$4$ & $3.001 \pm 0.495$ & $18.3$ & $14.594 \pm 56.745$ & $527$ & $13.159 \pm 2.459$ & $235$ \\ 
		$8$ & $5.617 \pm 0.542$ & $17.9$ & $15.242 \pm 58.458$ & $509$ & $14.781 \pm 3.501$ & $246$ \\
		$16$ & $11.218 \pm 0.722$ & $18.7$ & $14.466 \pm 45.581$ & $520$ & $14.765 \pm 2.480$ & $237$ \\ 
		$32$ & $22.575 \pm 1.269$ & $17.6$ & $16.819 \pm 46.574$ & $533$ & $18.216 \pm 3.069$ & $244$ \\ 
		$64$ & $45.047 \pm 2.538$ & $18.8$ & $22.288 \pm 46.398$ & $540$ & $25.403 \pm 4.393$ & $262$ \\
		\hline
	\end{tabular}
\end{table}

\clearpage
\section{Experiment with Energy Drain on Windows OS}
\label{app:energy-eff}

\begin{table*}[h!]
\footnotesize
\centering
\caption{Power drain over 200K iterations with a small dynamically constructed compute graph (Figure~\ref{fig:exp2-small-compute-graph}) consisting of 32 nodes, using FP64. Voltage: 11.7V, Battery: DELL J8FK941J, Chemistry: Lithium-ion polymer, OS: Windows 11, 1 mWh = 3.6 Joules.
See Figure~\ref{fig:exp2-power-drains-windows}. 
The numerical results across frameworks match exactly.
}

\begin{tabular}{|l|l|l|l|l|l|}
	\hline
	\textbf{{\footnotesize{\#}}} & 
	\textbf{{\footnotesize{\makecell[l]{Framework, Mode,\\Language, Device}}}} &
	\textbf{\parbox{2.0cm}{\center \footnotesize{Time with All Initialization (sec.)}}} & 
	\textbf{\parbox{2.0cm}{\center \footnotesize{Total Energy Consumed (mWh)}}} & 
	\textbf{\parbox{2.0cm}{\center \footnotesize{Task Energy Consumed (mWh)}}} & 
	\textbf{\parbox{2.0cm}{\center \footnotesize{OS Energy Consumed (mWh)}}} \\
	\hline
	\hline
	
	\cellcolor{bgcolorwe}1& 
	\cellcolor{bgcolorwe}\makecell[l]{BurTorch, Eager, \,\,\,\,\,\,\,\,\,\,\,\,\,\,\,\,\,\,\,\,\,\\C++, CPU} & 
	\cellcolor{bgcolorwe}0.089
	&\cellcolor{bgcolorwe}0.94& \cellcolor{bgcolorwe}0.593 & \cellcolor{bgcolorwe}0.347 \\
	\hline
	
	2&
	\makecell[l]{TensorFlow, Eager,\\Python, CPU}	& 239.321
	& 1710 & 776.64 & 933.35\\
	\hline
	
	3&
	\makecell[l]{TensorFlow, Graph,\\Semi-Python, CPU}	& 36.693
	& 293 & 149.89 & 143.10 \\
	\hline
	
	4&
	\makecell[l]{TF Lite, Graph,\\TF Lite Interpreter, CPU}&
	29.736
	& 304 & 188.03 & 115.97\\
	\hline
	
	5&
	\makecell[l]{Autograd, Eager,\\Python, CPU}& 
	381.295
	& 2901 & 1413.95 & 1487.05 \\
	\hline
	
	6&
	\makecell[l]{PyTorch, Eager,\\Python, \textbf{GPU}} & 
	427.265
	& 6014 
	& 4347.67
	& 1666.33 \\
	\hline
	
	7&
	\makecell[l]{PyTorch, Eager, \\Python, CPU} & 
	53.648 &
	408
	& 198.78 & 
	209.22 \\
	\hline
	
	8&
	\makecell[l]{PyTorch, Graph, \\TorchScript, CPU} & 
	53.870 &
	432
	& 221.90 & 210.09\\
	\hline
	
	9&
	\makecell[l]{PyTorch, Eager, \\LibTorch, C++, CPU}&
	31.300 & 
	280
	& 157.93 & 122.07\\
	\hline
	
	10&
	\makecell[l]{JAX, Eager, \\Python, CPU} &
	1794.833 & 14765 & 7765.16 & 6999.84\\
	\hline
	
	11&
	\makecell[l]{JAX, Graph, \\Semi-Python, CPU} &
	11.805 & 
	82
	& 35.97 & 46.03\\
	\hline
	
	12&
	\makecell[l]{Micrograd, Eager, \\Python, CPU}&
	10.691 & 
	71
	& 29.307 & 41.694
	\\
	\hline
	
	\hline
\end{tabular}
\label{tab:exp2-power-drains-windows}
\end{table*}

\begin{figure*}[ht]
\centering
\includegraphics[width=0.85\linewidth]{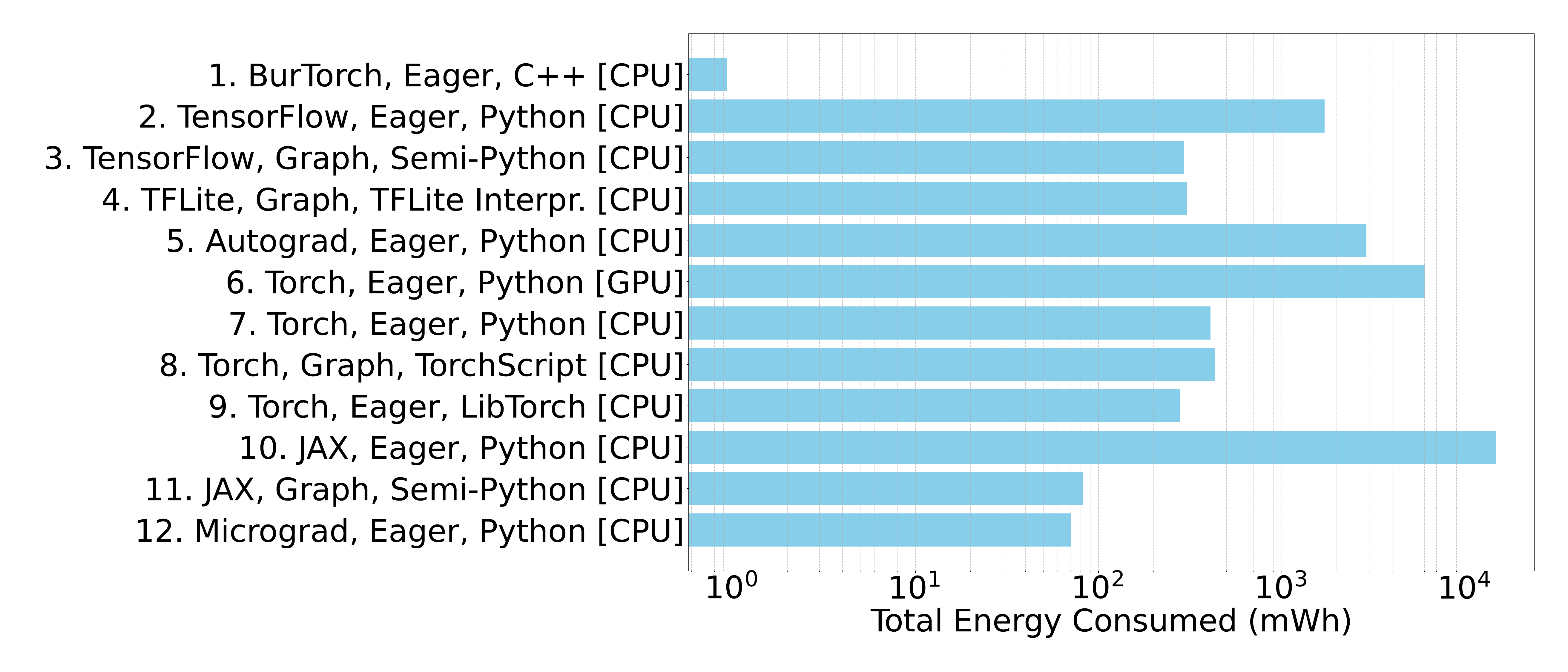}

\caption{Visualization of Table~\ref{tab:exp2-power-drains-windows}. Total power drain over 200K iterations with a \textit{small} dynamically constructed compute graph (Figure~\ref{fig:exp2-small-compute-graph}) consisting of 32 nodes, using FP64. Voltage: 11.7V, Battery: DELL J8FK941J, Chemistry: Li-poly, OS: Windows 11. The numerical results across frameworks match exactly.
}
\label{fig:exp2-power-drains-windows}
\end{figure*}

To evaluate energy consumption, we executed 200K iterations of the backpropagation algorithm across various computational frameworks for the computation graph from Figure~\ref{fig:exp2-small-compute-graph}. We compared end-to-end power consumption and end-to-end time. 

\paragraph{Preparation.} To ensure fair experimentation, we fixed all executions to run on physical CPU core number 2 (using 1-based indexing). The end-to-end computation was conducted on a Windows OS workstation, with the software and hardware environment detailed in Appendix~\ref{app:exp-setup-win}. The OS has 454 installed hardware drivers and 232 active processes. The high number of installed drivers is due to the presence of multiple hardware components. Having 232 running processes in Windows is standard practice, as many of them remain idle most of the time. Our system is equipped with the DELL J8FK941J battery. Charged Capacity: 92 828 mWh, Voltage: 11.102 volts, Chemistry: Lithium-ion polymer.

\paragraph{Cold state.}
Even when no applications are running, Windows OS core, installed hardware drivers, and system software impose a baseline load on the CPU. In this idle state, the cumulative CPU load ranges from 0.42\% to 0.76\% of its total processing capacity. During this cold phase, the system drains 3.9 mWh (milliwatt-hours) of energy per second from the battery. 
Thus, in the cold state, the hardware and OS software components consume power:
$$\mathrm{Consumed\,Power\,in\,Cold\,State} = 3.9 \cdot \dfrac{3600}{1000}=14.04 \mathrm{\,Joules\,per\,second}=14.04\mathrm{W}.$$

\paragraph{Measurements.}  
We use specific software frameworks to perform 200K iterations of backpropagation. In typical Deep Learning training, if 1000 epochs are used, with each epoch processing 200 data points, the total number of gradient oracle computations is approximately 200K. While this depends on the specific setup, the properties of \( f(x; D) \), the characteristics of the data \( D \), and the training algorithm, 200K iterations provide a reasonable (although rough) estimate.

We measure both time (in seconds) and energy (in mWh, where 1 mWh = 3.6 Joules)\footnote{To measure the battery capacity, we used the specialized tool BatteryInfoView v1.25 by NirSoft, which provides this information at one-second intervals.}. The execution includes time and energy for computation and memory transfers during end-to-end testing, covering all internal processes from initiating (for example, Python) to terminating the task in Windows 11, along with initialization, dynamic library loading, thread creation, and other overheads. The experiments run on a single $4.48$ GHz CPU core, with all hardware and software components powered solely by the installed battery. For more information about the environment, see Appendix~\ref{app:exp-setup-win}.

\paragraph{Results.}

The results are presented in Table~\ref{tab:exp2-power-drains-windows}. We observe (example 2) that, under light load, it is crucial to complete tasks quickly. As long as the OS is involved, it continues to drain energy. For example, in \libname{TensorFlow} Eager Mode experiment, the OS consumes more energy during the simulation than the application being executed. Next, we observe that using the GPU (examples 6 and 7) for small compute graphs not only increases wall-clock time but also leads to higher energy consumption. In general, non-Python solutions take less time and consume less energy (examples 2 and 3). However, relying solely on a compile-based environment (example 9) is insufficient. The choice of programming language plays a crucial role. \citet{pereira2021ranking} ranked languages based on various computational tasks, though not specifically for DL problems. Our work highlights that minimizing runtime overhead is equally critical, and this cannot be achieved solely by switching the programming language. Inefficiencies can arise in various forms in both scripting and compile-based language implementations. However, in scripting languages, when efficiency is paramount, the available tools are extremely limited.

Finally, as shown in Table~\ref{tab:exp2-power-drains-windows}, the energy consumed by \libname{BurTorch} is several orders of magnitude lower compared to current best-practice solutions. Specifically, \libname{BurTorch} (example 1) consumes only 0.94 mWh of total energy for 200K iterations, with 0.593 mWh dedicated to the task itself and 0.347 mWh for OS overhead. In contrast, \libname{TensorFlow} in Eager mode on CPU (example 2) consumes 1710 mWh, and \libname{PyTorch} in Eager mode on CPU (example 7) requires 408 mWh, highlighting a stark reduction in energy consumption by \libname{BurTorch}. These results demonstrate the significant efficiency advantages of \libname{BurTorch}, particularly when considering the energy demands of deploying this Deep Learning framework.


\clearpage
\section{Acknowledgments}

The research reported in this publication was supported by funding from King Abdullah University of Science and Technology (KAUST): i) KAUST Baseline Research Scheme, ii) Center of Excellence for Generative AI, under award number 5940.

In addition, we would like to acknowledge Ivan Ilin, a member of Peter Richt\'{a}rik’s Optimization and Machine Learning Laboratory, for his useful and insightful discussions.


\clearpage
\section{Limitations and Future Research}
\label{app:limitations}

While \libname{BurTorch} demonstrates significant improvements in computation speed, memory, and energy efficiency, several limitations should be acknowledged:

\begin{enumerate}
\item \textbf{Limited scalability to large batch sizes.} While \libname{BurTorch} outperforms \libname{PyTorch} in low-latency scenarios and small batch sizes, its advantages diminish as batch size increases. \libname{PyTorch's} optimized batching strategies and extensive parallelism enable better absolute performance for large-scale training workloads.

\item \textbf{CPU-centric design.}
\libname{BurTorch} is optimized for CPU execution and is not intended for GPU-accelerated environments. The organization of latency-aware GPU computations differs significantly from CPU-based execution, and exploring this difference remains an open question for future research.

\item \textbf{Optimization for small compute graphs.} \libname{BurTorch}'s efficiency gains are particularly suited for small compute graphs and latency-sensitive applications. Scaling \libname{BurTorch} to support larger graphs remains an open research question.

\item \textbf{Community and maintainability.} \libname{PyTorch} and \libname{TensorFlow} benefit from large developer communities and continuous updates, ensuring ongoing improvements and bug fixes. \libname{BurTorch}, in contrast, offers unit tests, a small codebase, and example implementations, but its development is expected to be driven primarily by external researchers. Establishing a feedback loop with contributors potentially through a plugin system is an area for future consideration.

\end{enumerate}

Despite these limitations, \libname{BurTorch} makes a strong case for scenarios where low-latency execution, as well as memory and energy efficiency, are crucial for CPU-based workloads.

\end{document}